\pdfoutput=1

\documentclass[sigconf, screen]{acmart}

\usepackage{bm}
\usepackage{cleveref}
\usepackage{multirow}
\usepackage{booktabs}
\usepackage[table]{xcolor}
\definecolor{highlightgray}{gray}{0.9}
\usepackage{graphicx}
\usepackage{algorithm}
\usepackage{algorithmic}

\usepackage{balance}
\usepackage{booktabs}
\usepackage{tabularx}
\usepackage{makecell}

\usepackage{booktabs}
\usepackage{siunitx}
\usepackage{threeparttable}

\usepackage[most,listings]{tcolorbox}
\usepackage{upquote}

\tcbuselibrary{listings}

\newtcblisting{promptbox}[2][]{%
  enhanced,
  breakable,
  boxed title style={
    colback=black!75,
    colframe=black!75},
  fonttitle=\footnotesize\bfseries,
  title={#2},
  listing only,
  listing options={
    basicstyle=\scriptsize\ttfamily,
    upquote=true,
    breaklines=true,
    breakautoindent=false,
    breakindent=0pt,
    showstringspaces=false,
    tabsize=2,
    columns=fullflexible},
  #1
}

\AtBeginDocument{%
  }

\setcopyright{acmlicensed}
\copyrightyear{2026}
\acmYear{2026}
\acmDOI{XXXXXXX.XXXXXXX}
\acmConference[Conference acronym 'XX]{Make sure to enter the correct
  conference title from your rights confirmation email}{June 03--05,
  2026}{Woodstock, NY}
\acmISBN{978-1-4503-XXXX-X/2018/06}

\acmSubmissionID{4982}

\begin{document}

\title{EVE: Verifiable Self-Evolution of MLLMs via Executable Visual Transformations}

\author{Yongrui Heng}
\authornote{These authors contributed equally to this work.}
\affiliation{%
  \institution{National Engineering Research Center for Software Engineering, Peking University}
  \city{Beijing}
  \country{China}
}
\email{1194913898@qq.com}

\author{Chaoya Jiang}
\authornotemark[1]
\authornote{Corresponding authors.}
\affiliation{%
  \institution{School of Control Science and Engineering, Shandong University}
  \city{Jinan}
  \country{China}
}
\email{jcy@sdu.edu.cn}

\author{Han Yang}
\affiliation{%
  \institution{Zeekr, Geely Auto}
  \city{Shanghai}
  \country{China}
}
\email{han.yang6@geely.com}

\author{Shikun Zhang}
\affiliation{%
  \institution{National Engineering Research Center for Software Engineering, Peking University}
  \city{Beijing}
  \country{China}
}
\email{zhangsk@pku.edu.cn}

\author{Wei Ye}
\authornotemark[2]
\affiliation{%
  \institution{National Engineering Research Center for Software Engineering, Peking University}
  \city{Beijing}
  \country{China}
}
\email{wye@pku.edu.cn}

\renewcommand{\shortauthors}{Yongrui Heng et al.}


\begin{abstract}
Self-evolution of multimodal large language models (MLLMs) remains a critical challenge: pseudo-label-based methods suffer from progressive quality degradation as model predictions drift, while template-based methods are confined to a static set of transformations that cannot adapt in difficulty or diversity. We contend that robust, continuous self-improvement requires not only deterministic external feedback independent of the model's internal certainty, but also a mechanism to perpetually diversify the training distribution. To this end, we introduce \textbf{EVE} (\textbf{E}xecutable \textbf{V}isual transformation based self-\textbf{E}volution), a novel framework that entirely bypasses pseudo-labels by harnessing executable visual transformations continuously enriched in both variety and complexity. EVE adopts a Challenger-Solver dual-policy architecture. The Challenger maintains and progressively expands a queue of visual transformation code examples, from which it synthesizes novel Python scripts to perform dynamic visual transformations. Executing these scripts yields VQA problems with absolute, execution-verified ground-truth answers, eliminating any reliance on model-generated supervision. A multi-dimensional reward system—integrating semantic diversity and dynamic difficulty calibration—steers the Challenger to enrich its code example queue while posing progressively more challenging tasks, preventing mode collapse and fostering reciprocal co-evolution between the two policies. Extensive experiments demonstrate that EVE consistently surpasses existing self-evolution methods, establishing a robust and scalable paradigm for verifiable MLLM self-evolution. The code is available at \url{https://github.com/0001Henry/EVE}.
\end{abstract}

\ccsdesc[500]{Computing methodologies~Artificial intelligence}
\keywords{Multimodal Large Language Models, Self-Evolution, Reinforcement Learning}




\maketitle

\section{Introduction}

The ability of a model to improve itself without human supervision has become a central goal in machine learning. For multimodal large language models (MLLMs), self-evolution refers to iteratively generating training data from the model's own capabilities and using that data to further improve the model—ideally without any human-labeled supervision or access to stronger external models. Recent efforts have made meaningful progress in this direction, but two fundamental problems persist.

\begin{figure}
    \centering
    \includegraphics[width=0.95\linewidth]{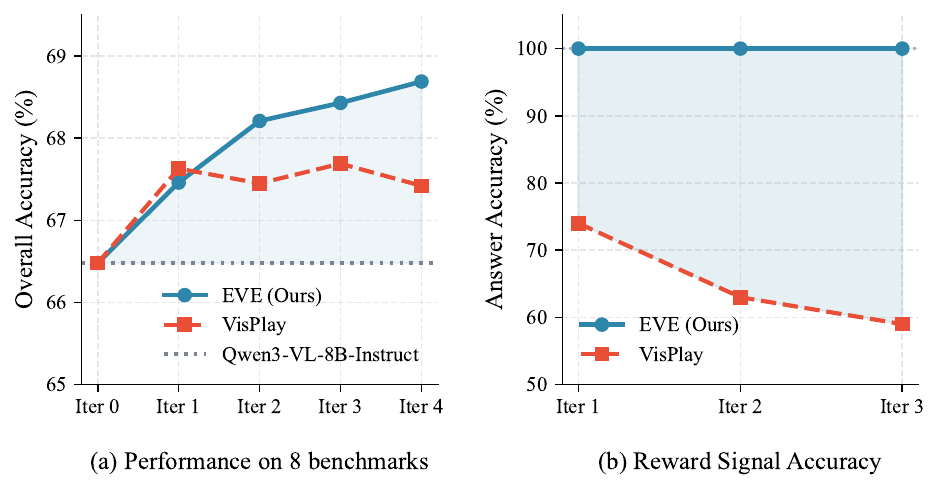}
    \vspace{-2ex}
    \caption{Quantitative comparison with VisPlay. (a) Overall accuracy on 8 benchmarks across iterations. (b) Reward signal accuracy (answer labels vs.\ ground truth) across iterations.}
    \label{fig:main_comparison}
    \vspace{-6ex}
\end{figure}

Pseudo-label degradation. Methods such as VisPlay~\cite{he2025visplay}, MM-Zero~\cite{li2026mm}, and EvoLMM~\cite{thawakar2025evolmm} construct training data by having the model make predictions and treating high-confidence outputs as pseudo-labels. This creates a positive feedback loop that works early in training but degrades rapidly later: as the model's predictions shift, pseudo-label quality degrades and error accumulates, as shown in Figure~\ref{fig:main_comparison}(b). Confidence scores do not measure ground-truth correctness.  
Recent theoretical analyses~\cite{he2026far,liu2026self} confirm that unsupervised reinforcement learning based on model confidence or entropy cannot reliably scale MLLM training. Furthermore, while difficulty filtering at intermediate success rates (around 50\%) has been theoretically justified for maximizing learning efficiency~\cite{bae2025online}, this result critically assumes accurate training labels. Pseudo-label-based methods violate this assumption, undermining their theoretical foundation and explaining their performance degradation over iterations.

Template diversity and difficulty limitations. An alternative to pseudo-labels is the proxy task paradigm, which constructs training questions from unlabeled images using predefined transformation templates. Agentic Jigsaw~\cite{zeng2025agentic} and JigsawR1~\cite{wang2025jigsawr1} use the jigsaw puzzle task as a proxy, where the model needs to reorder image patches to reconstruct the original image. PuzzleCraft~\cite{jeddi2025puzzle} extends this with rotation and patch-fitting. These approaches offer verifiable answers without human labels, but they are constrained to fixed, human-designed transformation templates. The question types, visual complexity, and difficulty levels are determined by the template designer, not by the model or the data, limiting the model's generalization capability to explore complex visual reasoning.

\begin{table}
    \caption{Comparison of human-annotation-free MLLM training methods. EVE is an external environment-based approach that simultaneously achieves accurate supervision, expandable question types, and no reliance on external models.}
    \vspace{-2ex}
    \label{tab:method_compare}
    \small
    \setlength{\tabcolsep}{1 pt}    
    \begin{tabular}{cccc}
        \toprule
        \textbf{Method} &
        \makecell{\textbf{Accurate}\\\textbf{Supervision}} &
        \makecell{\textbf{Expandable}\\\textbf{Questions}} &
        \makecell{\textbf{No External}\\\textbf{Model}} \\
        \midrule
        Pseudo-label-based~\cite{he2025visplay,li2026mm,thawakar2025evolmm}
            & no & yes & yes/no \\
        \midrule
        Template-based~\cite{zeng2025agentic,wang2025jigsawr1,jeddi2025puzzle} 
            & yes & no & yes \\
        \midrule
        EVE (Ours) & yes & yes & yes \\
        \bottomrule
    \end{tabular}
\vspace{-2ex}
\end{table}

Executable visual transformations as an environment. We argue that achieving expandable self-evolution requires a mechanism that (1) automatically generates diverse, image-adaptive questions, (2) provides accurate and verifiable correctness signals without human labels, and (3) adaptively controls task difficulty. Satisfying all three conditions requires the model to \textbf{learn from an external environment}—one where deterministic feedback is grounded in the environment's dynamics rather than the model's own predictions.

We identify \textbf{executable visual transformations}—Python programs that edit images—as a natural realization of such an environment. A Python function that transforms an image defines an infinite variety of visual puzzles whose ground-truth answers are determined entirely by program execution, not by the model. The space of possible transformations is as rich as the space of programs, and the resulting questions are as diverse as the images and parameters fed to those programs. Crucially, code execution provides an external, deterministic oracle: every question has a unique, verifiable answer independent of model confidence, resolving the pseudo-label degradation problem at its root. Moreover, because the model itself generates the transformation programs, the question space is expandable and adaptive—unconstrained by the several fixed templates—thereby overcoming the template diversity limitation.

Based on this insight, we propose \textbf{EVE} (\textbf{E}xecutable \textbf{V}isual transformation based self-\textbf{E}volution), a MLLM self-evolution framework built on a Challenger-Solver dual-policy architecture. Two policies are initialized from the same pretrained MLLM:
\begin{itemize}
    \item The \textbf{Challenger} receives an image and generates a Python program that defines an image editing function together with multiple candidate parameter sets. Executing the program produces the corresponding edited images.
    \item The \textbf{Solver} receives the original image, the code, the edited images, and a multiple-choice question, and tries to identify the correct parameter-to-image or image-to-parameter correspondence.
\end{itemize}

From each Challenger program, we automatically construct VQA questions with ground-truth labels determined entirely by code execution, independent of model confidence. The Challenger is guided by a multi-dimensional reward system that balances code validity, difficulty calibration (targeting $\sim$50\% Solver accuracy for optimal learning), and semantic diversity. To sustain the growth of question diversity, EVE maintains a visual transformation example queue that is progressively expanded with high-quality, semantically diverse code examples across iterations. At each iteration, the Challenger samples and composes codes from this queue to synthesize novel programs, ensuring that the question space continuously broadens rather than collapses to a narrow subset. The two policies co-evolve through alternating training, forming a self-sustaining loop of mutual improvement.

Our main contributions are:

\begin{itemize}
    \item We propose EVE, a self-evolution framework grounded in executable visual transformations as a programmatic external environment, where code execution provides verifiable ground-truth answers independent of model confidence, directly eliminating pseudo-label degradation.
    \item We design a Challenger-Solver dual-policy architecture with multi-dimensional rewards that enable adaptive question generation, overcoming the diversity and difficulty rigidity of template-based methods.
    \item Comprehensive experiments on 8 benchmarks demonstrate consistent improvements across both 4B and 8B model scales, validating that EVE continuously enhances visual perception and reasoning abilities.
\end{itemize}

\section{Related Work}
\label{sec:related}

\paragraph{Self-Evolution of Large Language Models.}
Self-evolution aims to improve model capabilities without human supervision or external models~\cite{deng2025self}. In language models, this paradigm has been explored through reinforcement learning: Absolute Zero~\cite{zhao2025absolute} uses code execution as a verifiable reward signal for unsupervised RL on coding and math tasks, while R-Zero~\cite{huang2025rzero} adopts a Challenger-Solver framework where majority vote serves as pseudo-labels. These approaches demonstrate the potential of self-evolution, but remain limited to text-only reasoning.

\begin{figure*}[t!]
\centering
  \includegraphics[width=1\textwidth]{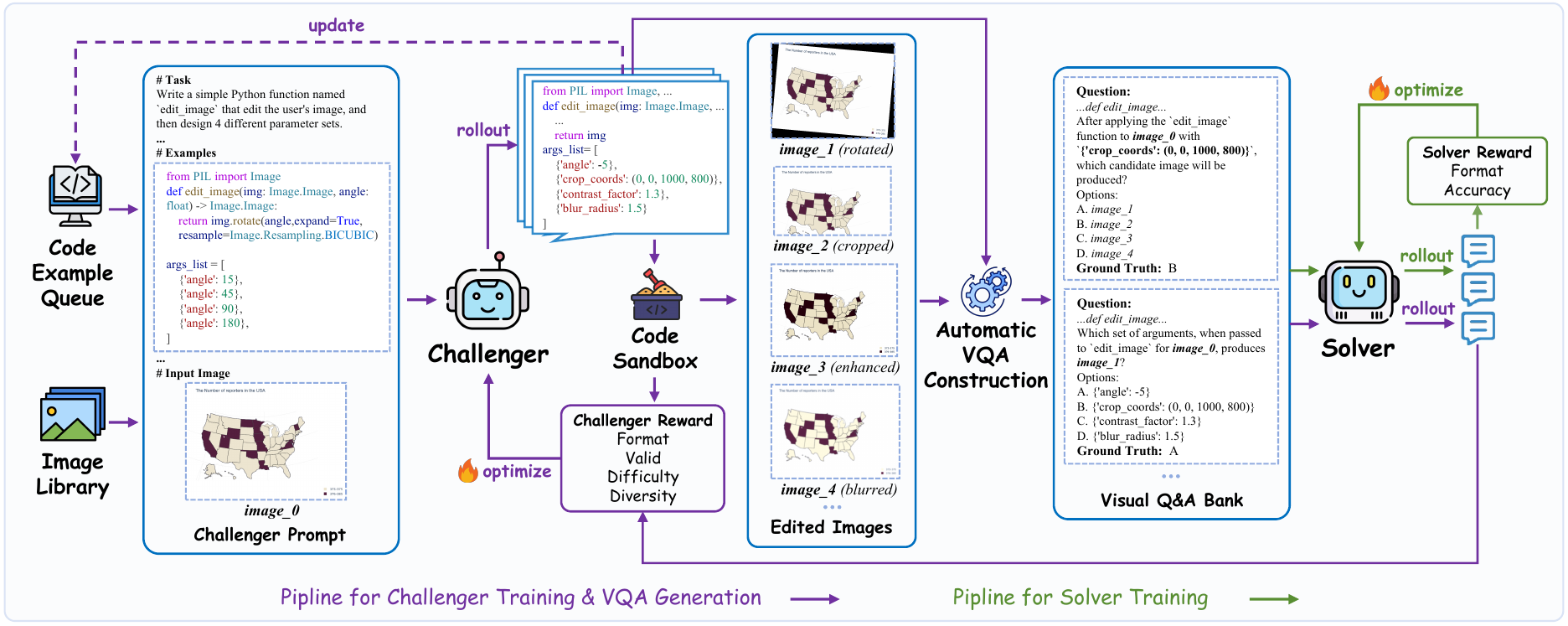}
  \vspace{-4ex}
  \caption{Overview of EVE. The model learns from executable visual transformations as a programmatic external environment: the Challenger generates Python scripts defining various visual transformations; script execution produces edited images and determines execution-verified ground-truth answers; VQA questions are automatically constructed to train the Solver; the Solver's accuracy feeds back as a difficulty signal to the Challenger. All supervision is derived from deterministic code execution, eliminating reliance on pseudo-labels or fixed templates.}
  \label{fig:teaser}
  \vspace{-3ex}
\end{figure*}

\paragraph{Self-Evolution of Multimodal Large Language Models.} For multimodal models, existing methods can be categorized into two paradigms, as shown in Table~\ref{tab:method_compare}: pseudo-label-based and template-based approaches.
\subparagraph{\textbf{Pseudo-label-based methods.}} These methods generate training data by having the model make predictions and treating certain outputs as supervision signals. This category can be further divided into two subcategories based on the source of supervision:
(1) \textit{Internal signal guided methods}~\cite{he2025visplay,li2026mm,thawakar2025evolmm,wang2026v,liu2024diving,sunil2026ireasoner} rely on the model's own confidence or entropy as supervision. VisPlay~\cite{he2025visplay} follows R-Zero's framework, constructing VQA pairs from unlabeled images and filtering by confidence scores. EvoLMM~\cite{thawakar2025evolmm} introduces continuous reward signals based on model certainty, while MM-Zero~\cite{li2026mm} extends this to settings without real images. However, confidence scores measure internal certainty rather than ground-truth correctness, causing error accumulation across iterations~\cite{he2025visplay,he2026far}.
(2) \textit{External model guided methods}~\cite{zhang2025viper,chen2025learning,khan2024dataenvgym} leverage external models as specialized verifiers for supervision. ViPER~\cite{zhang2025viper} uses diffusion models to externalize reasoning into visual snapshots, enabling the model to critique and refine its understanding. RRVF~\cite{chen2025learning} employs CLIP~\cite{radford2021learning} similarity as a reward signal, but remains limited to image-to-code tasks. While these methods can provide more reliable signals than internal confidence, they require access to proprietary models or domain-specific verifiers, limiting generalizability and violating the self-evolution principle.
\subparagraph{\textbf{Template-based methods.}} These methods~\cite{zeng2025agentic,wang2025jigsawr1,jeddi2025puzzle,liu2025spatial} construct proxy tasks with deterministic answers using predefined transformation templates. Agentic Jigsaw~\cite{zeng2025agentic} and JigsawR1~\cite{wang2025jigsawr1} shuffle image patches and ask models to predict the correct arrangement. Puzzle Curriculum GRPO~\cite{jeddi2025puzzle} extends this with rotation and patch-fitting tasks, introducing curriculum learning to gradually increase difficulty. Spatial-SSRL~\cite{liu2025spatial} is a similar paradigm that derives verifiable signals directly from ordinary RGB or RGB-D images. While these methods provide accurate supervision without pseudo-label degradation, they are constrained by fixed human-designed transformation templates. The question types, visual complexity, and difficulty levels are predetermined by template designers, preventing adaptive difficulty control and limiting the diversity of training data.

\paragraph{Code Generation for Vision-Language Tasks.}
Early work~\cite{suris2023vipergpt,liang2023code,guan2026codepercept} has explored using code as an intermediate representation for vision-language reasoning. Some recent "think with images" methods~\cite{hong2025deepeyesv2,zhao2025pyvision,zhang2025thyme,zhao2026pyvision,song2025codedance} use code as a tool to process the input image during the reasoning process, enabling more accurate and interpretable reasoning. However, these methods focus on inference-time program synthesis rather than using code execution as a training signal. In contrast, EVE leverages code execution as a self-evolution mechanism, providing deterministic supervision grounded in a programmatic external environment.
\section{Method}
\label{sec:method}

\subsection{Overview}

EVE addresses two core challenges in MLLM self-evolution: pseudo-label degradation and the diversity and difficulty rigidity of template-based proxy tasks, through a unified framework (Figure~\ref{fig:teaser}). The framework consists of two policies initialized from the same pretrained MLLM: a \textbf{Challenger} policy $\pi_C$ that generates executable Python code defining visual transformations, and a \textbf{Solver} policy $\pi_S$ that answers visual questions constructed from code execution results.

Given an image library $\mathcal{D}$, the Challenger samples an image $I \in \mathcal{D}$ and generates code $P$ that produces multiple edited images. From these transformations, we automatically construct visual questions $q$ with verifiable ground-truth answers $y^*$ determined by code execution. The Solver then attempts to answer these questions.

The two policies co-evolve through alternating optimization over $T$ iterations: the Challenger learns to generate diverse, moderately difficult questions that challenge the Solver, while the Solver learns to answer these questions correctly. This adversarial-cooperative dynamic drives continuous improvement without external supervision.

\subsection{Challenger Policy}
\paragraph{Task Definition.}
Given an input image $I$, the Challenger generates executable Python code $P$ containing: (1) an \texttt{edit\_image(img, **args)} function that applies one or more transformations to the image, and (2) a list \texttt{args\_list} of $N$ distinct parameter sets $\mathcal{A} = \{a_1, \ldots, a_N\}$. Executing the code produces $N$ edited images $\{I_1, \ldots, I_N\}$ where $I_j = \texttt{edit\_image}(I, a_j)$ for $j \in \{1, \ldots, N\}$.

\paragraph{Code Generation with Few-Shot Examples.}
To guide the Challenger, we maintain a priority queue $\mathcal{Q}$ of high-quality code examples. The priority queue is initialized with four seed examples covering basic transformations (jigsaw puzzles, rotation, cropping, bounding box drawing) and evolves during training to accumulate diverse, high-reward programs.

For each generation, we randomly sample $N_{\text{e}}$ examples from $\mathcal{Q}$ and construct a prompt that includes: (1) task instructions and requirements, (2) the example codes, and (3) the input image $I$. The Challenger is prompted to output code in the Python code block.

\paragraph{Execution-verified task synthesis.}
From the generated code $P$ and execution results $\{I_j\}_{j=1}^N$, we automatically construct two types of multiple-choice questions:

\subparagraph{\textbf{Type-0 (Parameter-to-Image).}} Given the original image $I$, the code $P$, one parameter set $a_j \in \mathcal{A}$, and $N$ candidate images $\{I_1, \ldots, I_N\}$, identify which image corresponds to $a_j$. This tests whether the Solver can execute the code mentally and predict the visual outcome.

\subparagraph{\textbf{Type-1 (Image-to-Parameter).}} Given the original image $I$, one transformed image $I_j$, the code $P$, and $N$ parameter options $\mathcal{A} = \{a_1, \ldots, a_N\}$, identify which parameters produced $I_j$. This tests whether the Solver can perform inverse reasoning from visual observations to parameter values.

Both question types have unique, verifiable answers determined entirely by code execution, eliminating reliance on model confidence or human judgment.
We apply random shuffling to answer options while maintaining the one-to-one correspondence between parameters and their resulting images. This ensures the model must attend to visual content rather than exploiting positional biases.

\paragraph{Challenger Reward Function.}
The Challenger's total reward combines four components:
\begin{equation}
    r_C = \lambda_{\text{format}} r_{\text{format}} + \lambda_{\text{valid}} r_{\text{valid}} + \lambda_{\text{diff}} r_{\text{diff}} + \lambda_{\text{div}} r_{\text{div}}
\label{eq:challenger_reward}
\end{equation}

\subparagraph{\textbf{Format Reward}} $r_{\text{format}} \in \{0,1\}$ encourages parseable code blocks and penalizes irregular formatting or comments. To avoid solvers taking shortcuts from comments or being misleading, we require Challenger to generate uncommented code.

\subparagraph{\textbf{Validity Reward}} $r_{\text{valid}} \in \{0,1\}$ requires successful sandbox execution with constraints: no random functions, no duplicate parameters in \texttt{args\_list}, and image quality filtering (size bounds, no identical outputs). 
Specifically, we compute the similarity between the edited images through perceptual hashing, and if the similarity exceeds a threshold of 0.95, the output is considered to be a duplicate.

\subparagraph{\textbf{Difficulty Reward}} $r_{\text{diff}} \in [0,1]$ calibrates question difficulty. For each generated question $q$, we sample the Solver $K$ times to compute average accuracy:
\begin{equation}
\text{acc}_s(q) = \frac{1}{K}\sum_{k=1}^K \mathbb{1}[\hat{y}_k = y^*]
\label{eq:accuracy}
\end{equation}
where $\hat{y}_k \sim \pi_S(q)$ is the $k$-th sampled answer and $y^*$ is the ground truth. The difficulty reward is:
\begin{equation}
r_{\text{diff}}(q) = 1 - 2|\text{acc}_s(q) - 0.5|
\label{eq:difficulty}
\end{equation}
This function peaks at $\text{acc}_s = 0.5$ (moderately difficult) and decreases toward 0 or 1 (too easy or too hard), encouraging questions that maximize learning gain~\cite{bae2025online}.
The difficulty reward for code $P$ is computed as the average difficulty reward over all questions it generates.

\subparagraph{\textbf{Diversity Reward}} $r_{\text{div}} \in [-1,0]$ prevents mode collapse. For a batch of $G$ generated samples $\mathcal{X}_I = \{x_1, \ldots, x_G\}$ given image $I$, we cluster them by pairwise BLEU similarity~\cite{papineni2002bleu} into $K_I$ clusters $\{C_1^{(I)}, \ldots, C_{K_I}^{(I)}\}$. For sample $x_i$ in cluster $C_k^{(I)}$, the redundancy density is:
\begin{equation}
p_i = \frac{|C_k^{(I)}|}{G}
\end{equation}
After min-max normalization over the batch, the diversity reward is:
\begin{equation}
r_{\text{div}}(x_i) = -\frac{p_i - \min_{j} p_j}{\max_{j} p_j - \min_{j} p_j}
\end{equation}
This penalizes samples in large clusters, encouraging exploration of diverse programs.

\subsection{Solver Policy}

The Solver receives a question $q$ (including images, code, and options) and generates an answer $\hat{y}$. Its reward function is:
\begin{equation}
r_S = \omega_{\text{format}} r_{\text{format}} + \omega_{\text{acc}} r_{\text{acc}}
\end{equation}
The format reward $r_{\text{format}} \in \{0,1\}$ ensures the answer is extractable and properly formatted, while the accuracy reward $r_{\text{acc}} \in \{0,1\}$ measures semantic correctness against the ground-truth answer $y^*$ determined by code execution.

\vspace{-2ex}
\subsection{Self-Evolution Loop}

The Challenger and Solver co-evolve through alternating optimization over $T$ iterations. Algorithm~\ref{alg:self_evolution} presents the complete training procedure.

The priority queue $\mathcal{Q}$ maintains the top-$M$ code examples ranked by $r_{\text{diff}}$. For codes with equal $r_{\text{diff}}$, later additions have priority. BLEU-based deduplication ensures semantic diversity, codes exceeding $\sigma_{\text{high}}$ are discarded. This alternating optimization drives continuous improvement: the Challenger learns to generate diverse, moderately difficult questions, while the Solver learns to answer them correctly, creating a virtuous cycle of self-evolution.

\begin{algorithm}[t]
\caption{Self-Evolution Loop of EVE}
\label{alg:self_evolution}
\begin{algorithmic}[1]
\REQUIRE Initial policies $\pi_C^0$, $\pi_S^0$; Image library $\mathcal{D}$; Priority queue $\mathcal{Q}$; Total iterations $T$; Length of args\_list $N$; Number of in-context code examples $N_{\text{e}}$; Training steps per iteration $N_{\text{step}}$; Batch size $B$; Solver's sampling size $K$; 

\ENSURE Evolved policies $\pi_C^T$, $\pi_S^T$
\FOR{$t = 1$ to $T$}
    \STATE Initialize question bank $\mathcal{Y}_t \leftarrow \emptyset$
    \STATE \textcolor{gray}{// Phase 1: Train Challenger \& VQA generation}
    \FOR{$\text{step} = 1$ to $N_{\text{step}}$}
        \STATE Sample image $I \sim \mathcal{D}$ and $N_{\text{e}}$ examples $\{P_m\}_{m=1}^M \sim \mathcal{Q}$
        \STATE Generate code $P \sim \pi_C^t(I, \{P_m\}_{m=1}^M)$
        \STATE Execute $P$ to obtain edited images $\{I_j\}_{j=1}^N$
        \STATE Construct questions $q_0, q_1$ from $(I, P, \{I_j\}_{j=1}^N)$
        \STATE \textcolor{gray}{// Evaluate difficulty}
        \FOR{each question $q \in \{q_0, q_1\}$}
            \STATE Sample answers $\{\hat{y}_k\}_{k=1}^K \sim \pi_S^t(q)$
            \STATE Compute accuracy $\text{acc}_s(q)$ \textcolor{gray}{// see Eq.~\ref{eq:accuracy}}
            \STATE Compute difficulty reward $r_{\text{diff}}(q)$ \textcolor{gray}{// see Eq.~\ref{eq:difficulty}}
            \STATE Add $(q, r_{\text{diff}}(q))$ to question bank $\mathcal{Y}_t$
        \ENDFOR
        \STATE $\bar{r}_{\text{diff}} \leftarrow \frac{1}{2}(r_{\text{diff}}(q_0) + r_{\text{diff}}(q_1))$
        \STATE Add $(P, \bar{r}_{\text{diff}})$ to priority queue $\mathcal{Q}$
        \STATE $r_C \leftarrow \sum_{k} \lambda_k r_k$ \textcolor{gray}{// see Eq.~\ref{eq:challenger_reward}}
        \STATE Update $\pi_C^t$ with reward $r_C$ \textcolor{gray}{// see Eq.~\ref{eq:training_objective}}
    \ENDFOR
    \STATE \textcolor{gray}{// Phase 2: Train Solver}
    \STATE Sort $\mathcal{Y}_t$ by $r_{\text{diff}}$ in descending order
    \STATE $\mathcal{Y}_t \leftarrow \mathcal{Y}_t[: B \cdot N_{\text{step}}]$
    \STATE Train $\pi_S^t$ on question bank $\mathcal{Y}_t$ for $N_{\text{step}}$ steps with $r_S$
    \STATE Obtain updated Solver $\pi_S^{t+1}$
\ENDFOR
\RETURN $\pi_C^T$, $\pi_S^T$
\end{algorithmic}
\end{algorithm}
\vspace{-2ex}

\subsection{Training Objective}

Both the Challenger and Solver policies are optimized using Reinforcement Learning (RL). Following DAPO~\cite{yu2025dapo}, we apply global normalization averaging over all tokens while retaining the KL divergence regularization term from GRPO~\cite{shao2024deepseekmath}. The optimization objective is:
\begin{equation}
\begin{aligned}
J_{\text{RL}}(\theta) = \mathbb{E}_{\substack{q \sim \mathcal{D} \\ \{o_i\}_{i=1}^{G}\sim \pi_{\theta_{\text{old}}}(\cdot|q)}} \Bigg[ &\frac{1}{\sum_{i=1}^{G}|o_i|} \sum_{i=1}^{G}\sum_{t=1}^{|o_i|} \Big( \min\big( r_{i,t}(\theta)\hat A_{i,t}, \\
&\text{clip}(r_{i,t}(\theta), 1-\epsilon_{\text{low}}, 1+\epsilon_{\text{high}})\hat A_{i,t} \big) \\
&- \beta D_{\text{KL}}(\pi_\theta||\pi_{\text{ref}}) \Big) \Bigg]
\end{aligned}
\label{eq:training_objective}
\end{equation}
where $q$ is the input query (image and prompt for Challenger, or question for Solver), $\{o_i\}_{i=1}^{G}$ are $G$ sampled outputs, and $|o_i|$ denotes the token length of output $o_i$. The probability ratio is:
\begin{equation}
r_{i,t}(\theta) = \frac{\pi_\theta(o_{i,t}\mid q,o_{i,<t})}{\pi_{\theta_{\text{old}}}(o_{i,t}\mid q,o_{i,<t})}
\end{equation}
and the advantage function is computed as:
\begin{equation}
\hat A_{i,t} = \frac{R_i-\text{mean}(\{R_j\}_{j=1}^{G})}{\text{std}(\{R_j\}_{j=1}^{G})}
\end{equation}
where $R_i$ is the total reward for output $o_i$ (either $r_C$ for Challenger or $r_S$ for Solver). The advantage is shared across all tokens in the same output, encouraging coherent generation. The KL divergence penalty prevents the policy from deviating too far from the reference policy $\pi_{\text{ref}}$, maintaining stability during training.

\section{Experiments}
\label{sec:experiments}

\begin{table*}
\centering
\caption{Performance comparison of EVE with VisPlay on Qwen3-VL-8B over 3 iterations.}
\vspace{-2ex}
\label{tab:8b_iterations}
\small
\setlength{\tabcolsep}{3pt}
\begin{tabular}{lccccccccc}
\toprule
\multirow{2}{*}{\textbf{Method}} & \multicolumn{2}{c}{\textbf{General VQA}} & \multicolumn{2}{c}{\textbf{Alignment}} & \multicolumn{2}{c}{\textbf{Math Reasoning}} & \textbf{Perception} & \textbf{Multi-Image} & \multirow{2}{*}{\textbf{Overall}} \\
& \textbf{MMStar} & \textbf{MMVet} & \textbf{HalluBench} & \textbf{MIA-Bench} & \textbf{VisuLogic} & \textbf{MathVista} & \textbf{BLINK} & \textbf{Muirbench} & \\
\midrule
Qwen3-VL-8B-Instruct & 72.07 & 66.79 & 61.18 & 91.98 & 24.6 & 76.9 & 65.02 & 73.27 & 66.48 \\
\midrule
VisPlay-iter1 & 73.40 & 69.82 & 61.07 & 91.46 & 25.6 & 78.5 & \underline{65.83} & \textbf{75.38} & 67.63 \\
VisPlay-iter2 & 72.67 & \textbf{72.29} & 60.57 & 92.35 & 23.7 & 77.7 & 65.12 & \underline{75.23} & 67.45 \\
VisPlay-iter3 & 73.13 & 69.22 & 61.56 & 93.02 & \underline{26.3} & 78.8 & 64.49 & 74.96 & 67.69 \\
\midrule
Ours-iter1 & 73.47 & 70.00 & \underline{62.70} & 91.05 & 24.6 & \underline{79.2} & 65.44 & 73.23 & 67.46 \\
Ours-iter2 & \underline{74.20} & \underline{71.74} & 62.19 & \textbf{93.31} & 25.3 & \textbf{79.3} & 65.49 & 74.12 & \underline{68.21} \\
Ours-iter3 & \textbf{74.53} & 71.10 & \textbf{62.78} & \underline{92.99} & \textbf{27.2} & 77.8 & \textbf{66.81} & 74.23 & \textbf{68.43} \\
\bottomrule
\end{tabular}
\vspace{-1ex}
\end{table*}

\begin{table*}
\centering
\caption{Extended training results on Qwen3-VL-8B up to 5 iterations.}
\vspace{-2ex}
\label{tab:extended_iterations}
\small
\setlength{\tabcolsep}{3pt}
\begin{tabular}{lccccccccc}
\toprule
\multirow{2}{*}{\textbf{Method}} & \multicolumn{2}{c}{\textbf{General VQA}} & \multicolumn{2}{c}{\textbf{Alignment}} & \multicolumn{2}{c}{\textbf{Math Reasoning}} & \textbf{Perception} & \textbf{Multi-Image} & \multirow{2}{*}{\textbf{Overall}} \\
& \textbf{MMStar} & \textbf{MMVet} & \textbf{HalluBench} & \textbf{MIA-Bench} & \textbf{VisuLogic} & \textbf{MathVista} & \textbf{BLINK} & \textbf{Muirbench} & \\
\midrule
Qwen3-VL-8B-Instruct & 72.07 & 66.79 & 61.18 & 91.98 & 24.6 & 76.9 & 65.02 & 73.27 & 66.48 \\
\midrule
Ours-iter3 & \textbf{74.53} & \underline{71.10} & \underline{62.78} & 92.99 & 27.2 & 77.8 & 66.81 & 74.23 & 68.43 \\
Ours-iter4 & \underline{73.93} & \textbf{71.88} & \textbf{62.99} & \textbf{93.31} & \underline{27.8} & \underline{77.9} & \textbf{67.33} & \underline{74.38} & \textbf{68.69} \\
Ours-iter5 & 73.53 & 70.87 & \underline{62.78} & \underline{93.19} & \textbf{27.8} & \textbf{78.1} & \underline{67.18} & \textbf{76.00} & \underline{68.68} \\
\bottomrule
\end{tabular}
\vspace{-1ex}
\end{table*}

\begin{table*}
\centering
\caption{Comparison with state-of-the-art MLLMs and self-evolution methods.}
\vspace{-2ex}
\label{tab:sota-comparison}
\small
\setlength{\tabcolsep}{3pt}
\begin{tabular}{lccccccccc}
\toprule
\multirow{2}{*}{\textbf{Method}} & \multicolumn{2}{c}{\textbf{General VQA}} & \multicolumn{2}{c}{\textbf{Alignment}} & \multicolumn{2}{c}{\textbf{Math Reasoning}} & \textbf{Perception} & \textbf{Multi-Image} & \multirow{2}{*}{\textbf{Overall}} \\
& \textbf{MMStar} & \textbf{MMVet} & \textbf{HalluBench} & \textbf{MIA-Bench} & \textbf{VisuLogic} & \textbf{MathVista} & \textbf{BLINK} & \textbf{Muirbench} & \\
\midrule
\multicolumn{10}{c}{\textit{Closed-Source MLLMs}} \\
\midrule
GPT-5 mini (minimal) & 61.3 & - & - & 92.3 & \underline{27.6} & 59.6 & 56.7 & 57.5 & - \\
GPT-5 nano (high) & 68.6 & - & - & 89.9 & 24.5 & 71.5 & 58.3 & 65.7 & - \\
GPT-4o-20240513 & 64.7 & 69.1 & 55.0 & - & 26.3 & 63.8 & \textbf{68.0} & 68.0 & - \\
\midrule
\multicolumn{10}{c}{\textit{Open-Source MLLMs}} \\
\midrule
InternVL3-9B & 66.3 & \textbf{76.2} & 51.2 & - & - & 71.5 & 58.6 & 51.4 & - \\
LLaVA-OneVision-72B & 65.8 & 60.6 & 49.0 & - & - & 67.1 & 55.4 & 54.8 & - \\
Qwen3-VL-8B-Instruct & 72.1 & 66.8 & 61.2 & 92.0 & 24.6 & 76.9 & 65.0 & 73.3 & 66.5 \\
\midrule
\multicolumn{10}{c}{\textit{Template-based Self-evolution Methods}} \\
\midrule
Spatial-SSRL-7B & - & - & 53.2 & - & - & - & 56.2 & - & - \\
Jigsaw-R1-8B & \textbf{74.3} & 69.4 & \underline{62.1} & 91.3 & 27.4 & \underline{77.9} & 64.7 & 73.2 & 67.5 \\
\midrule
\multicolumn{10}{c}{\textit{Pseudo-label-based Self-evolution Methods}} \\
\midrule
Qwen-ViPER-7B & 66.2 & 65.8 & 54.4 & - & - & - & 57.6 & - & - \\
VisPlay-8B-iter3 & 73.1 & 69.2 & 61.6 & \underline{93.0} & 26.3 & \textbf{78.8} & 64.5 & \textbf{75.0} & \underline{67.7} \\
MM-Zero-8B-iter3 & 70.7 & 69.5 & 61.7 & 92.9 & 25.7 & 74.7 & \underline{65.9} & 72.2 & 66.7 \\
\midrule
Ours-8B-iter4 & \underline{73.9} & \underline{71.9} & \textbf{63.0} & \textbf{93.3} & \textbf{27.8} & \underline{77.9} & \underline{67.3} & \underline{74.4} & \textbf{68.7} \\
\bottomrule
\end{tabular}
\vspace{-2ex}
\end{table*}

\subsection{Experimental Setup}

\paragraph{Implementation Details.}
Both Challenger and Solver policies are initialized from the same base checkpoint.
For the Challenger, we set the length of args\_list $N=4$, the number of in-context code examples $N_{\text{e}}=2$, sampling rounds $K=6$ for difficulty estimation, and batch size $B=128$. The priority queue maintains the top-50 ($M=50$) code examples. We train for $T=3$ iterations with $N_{\text{step}}=10$ training steps per iteration. In the task synthesis phase, we randomly select a parameter set-edited image pair and construct one problem each for Type-0 and Type-1 categories. For reward weights, we set $\lambda_{\text{format}}=0.2$, $\lambda_{\text{valid}}=0.4$, $\lambda_{\text{diff}}=0.4$, $\lambda_{\text{div}}=0.3$ for the Challenger, and $\omega_{\text{format}}=0.2$, $\omega_{\text{acc}}=0.8$ for the Solver. The image library is derived from the Vision-SR1-47K dataset\cite{li2025selfrewardingvisionlanguagemodelreasoning}, which contains high-quality images from diverse domains.
More details are provided in the Appendix.

\paragraph{Baselines.}
We first instantiate EVE on Qwen3-VL-Instruct models~\cite{bai2025qwen3}.
We compare with pseudo-label-based self-evolution methods VisPlay~\cite{he2025visplay}, MM-Zero~\cite{li2026mm}, and ViPER~\cite{zhang2025viper}; and template-based methods Jigsaw-R1~\cite{wang2025jigsawr1} and Spatial-SSRL~\cite{liu2025spatial}. For fair comparison, we reproduce two representative methods, VisPlay and Jigsaw-R1 on Qwen3-VL-8B-Instruct using the same training hyperparameters and image corpus as our method.
We also compare against closed-source models GPT-5 mini/nano~\cite{singh2025openai} and GPT-4o~\cite{hurst2024gpt}; and larger open-source MLLMs InternVL3-9B~\cite{zhu2025internvl3} and LLaVA-OneVision-72B~\cite{li2024llava}.

\paragraph{Evaluation Setting.}
We adopt VLMEvalKit~\cite{duan2024vlmevalkit}, a widely used evaluation framework, for all assessments.
We evaluate performance across diverse benchmarks covering multiple capabilities: MMStar~\cite{chen2024we} and MMVet~\cite{yu2023mm} (general VQA); HallusionBench~\cite{guan2024hallusionbench} (visual hallucination); MIA-Bench~\cite{qian2024mia} (complex instruction following); VisuLogic~\cite{xu2025visulogic} and MathVista (testmini subset)~\cite{lu2023mathvista} (mathematical and logical reasoning); BLINK~\cite{fu2024blink}(visual perception); MuirBench~\cite{wang2024muirbench} (multi-image understanding).
In total, the evaluation suite comprises more than 10{,}000 test samples, comprehensively covering a wide variety of multimodal understanding and reasoning tasks.

\subsection{Main Results}

\paragraph{Comparison with VisPlay over 3 iterations.}
We first evaluate on Qwen3-VL-8B-Instruct and compare with the base model as well as VisPlay~\cite{he2025visplay}, a representative pseudo-label-based self-evolution method.
Results are shown in Table~\ref{tab:8b_iterations}.
After 3 iterations, EVE achieves an overall score of 68.43, outperforming both the base model (66.48, +1.95) and VisPlay-iter3 (67.69, +0.74).
On individual benchmarks, EVE at iteration 3 obtains the best results on MMStar (74.53), HallusionBench (62.78), VisuLogic (27.2), and BLINK (66.81), demonstrating strong improvements in comprehensive evaluation, hallucination resistance and visual logical reasoning.

\subparagraph{\textbf{Addressing pseudo-label degradation.}} A key distinction lies in the training dynamics. Our method exhibits monotonically increasing overall performance across iterations (67.46 → 68.21 → 68.43), whereas VisPlay fluctuates (67.63 → 67.45 → 67.69), with a notable drop at iteration 2. Furthermore, EVE maintains stable or improved performance across all individual benchmarks without any regression.
For instance, on HallusionBench, VisPlay suffers from performance fluctuations and even degradation, suggesting that pseudo-labels generated by the evolving model may become less reliable over time and further amplify hallucinations, whereas EVE mitigates hallucination by grounding learning in code execution feedback.
This empirically confirms that learning from accurate reward signals resolves the fundamental limitation of pseudo-label methods.

\paragraph{Extended Training Analysis.}
To analyze long-term evolution trends, we extend training up to 5 iterations, with results shown in Table~\ref{tab:extended_iterations} and Figure~\ref{fig:main_comparison}(a).
Our method achieves the highest overall score of 68.7, outperforming the base model by +2.2. The most substantial gains are observed on MMVet (+5.1), VisuLogic (+3.2), and BLINK (+2.3).
Performance stabilizes at iteration 5 (68.68), with some benchmarks exhibiting marginal trade-offs.
Crucially, the absence of significant performance collapse over 5 iterations demonstrates the stability of code-execution-based rewards, contrasting sharply with pseudo-label methods that typically degrade after 2--3 iterations.

\paragraph{Comparison with Existing Advanced MLLMs and Self-Evolution Methods.}
Table~\ref{tab:sota-comparison} compares our 8B model against recent advanced closed-source MLLMs, open-source models, and self-evolution baselines. For self-evolution methods, we report results from the best-performing iteration checkpoint.
Our method achieves the highest overall score of 68.7, surpassing all compared self-evolution approaches.
Relative to self-evolution baselines, EVE outperforms VisPlay (67.7) by +1.0 and MM-Zero (66.7) by +2.0. Particularly noteworthy is the performance on BLINK, where EVE achieves 67.3, exceeding both the base model (+2.3) and all other self-evolution methods, demonstrating substantial enhancement in visual perception capabilities.
While Jigsaw-R1 leads on MMStar and VisPlay leads on MathVista and MuirBench, EVE achieves the best performance on all remaining benchmarks, indicating broader and more balanced improvements.

\subparagraph{\textbf{Addressing template limitations.}} Jigsaw-R1's fixed templates improve MMStar and VisuLogic but constrain visual perception and multi-image understanding, leading to weaker performance on BLINK and MuirBench. By contrast, the Challenger autonomously discovers diverse editing operations through code generation, exploring richer visual reasoning patterns beyond fixed, human-designed transformation templates.


\subsection{Ablation Studies}

\begin{table}
\centering
\caption{Ablation study on challenger reward components.}
\label{tab:ablation}
\vspace{-2ex}
\begin{threeparttable}
\setlength{\tabcolsep}{1.5pt}
\footnotesize
\begin{tabular}{l
                S[table-format=2.2]
                S[table-format=2.2]
                S[table-format=2.1]
                S[table-format=2.2]
                S[table-format=2.2]}
\toprule
\textbf{Method} & \textbf{MMStar} & \textbf{HalluBench} & \textbf{MathVista} & \textbf{BLINK} & \textbf{Overall} \\
\midrule
Qwen3-VL-8B-Instruct & 72.07 & 61.18 & 76.90 & 65.02 & 68.79 \\
\midrule
\multicolumn{6}{l}{\textit{Full Method}} \\
\quad ours-iter1 & 73.47 & 62.70 & 79.20 & 65.44 & 70.20 \\
\quad ours-iter2 & 74.20 & 62.19 & 79.30 & 65.49 & 70.30 \\
\quad ours-iter3 & 74.53 & 62.78 & 77.80 & 66.81 & 70.48 \\
\midrule
\multicolumn{6}{l}{\textit{w/o $r_{\text{div}}$}} \\
\quad ours-iter1 & 73.80 & 61.53 & 77.30 & 63.23 & {68.97\,\textcolor{red}{\scriptsize($-$1.23)}} \\
\quad ours-iter2 & 74.47 & 62.38 & 78.10 & 63.18 & {69.53\,\textcolor{red}{\scriptsize($-$0.77)}} \\
\quad ours-iter3 & 73.93 & 61.16 & 77.70 & 64.80 & {69.40\,\textcolor{red}{\scriptsize($-$1.08)}} \\
\midrule
\multicolumn{6}{l}{\textit{w/o $r_{\text{div}}$ + w/o $r_{\text{diff}}$}} \\
\quad ours-iter1 & 73.20 & 61.36 & 77.10 & 64.14 & {68.95\,\textcolor{red}{\scriptsize($-$1.25)}} \\
\quad ours-iter2 & 73.87 & 61.28 & 77.40 & 63.02 & {68.89\,\textcolor{red}{\scriptsize($-$1.41)}} \\
\quad ours-iter3 & 72.73 & 61.03 & 76.70 & 63.80 & {68.57\,\textcolor{red}{\scriptsize($-$1.91)}} \\
\bottomrule
\end{tabular}
\end{threeparttable}
\vspace{-2ex}
\end{table}

\begin{figure}
    \centering
    \includegraphics[width=0.75\linewidth]{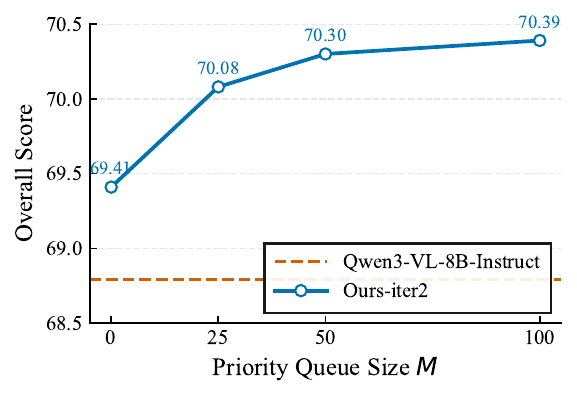}
    \vspace{-2ex}
    \caption{Ablation on priority queue size $M$. Overall score (average of MMStar, HallusionBench, MathVista, and BLINK) at iteration 2 under different $M$ values.}
    \label{fig:ablation_queue}
    \vspace{-2ex}
\end{figure}

\begin{table}
\centering
\caption{Results on Qwen3-VL-4B across 3 iterations.}
\vspace{-2ex}
\label{tab:4b_results}
\footnotesize
\setlength{\tabcolsep}{1pt}
\begin{tabular}{lcccccc}
\toprule
\textbf{Method} & \textbf{MMStar} & \textbf{MMVet} & \textbf{HalluBench} & \textbf{VisuLogic} & \textbf{BLINK} & \textbf{Overall} \\
\midrule
Qwen3-VL-4B-Instruct & 70.33 & 67.89 & 60.08 & 24.2 & 65.18 & 57.54 \\
\midrule
Ours-iter1 & \textbf{71.20} & \underline{68.99} & 59.98 & 26.1 & \underline{65.33} & 58.32 \\
Ours-iter2 & 70.53 & 68.58 & \textbf{61.82} & \underline{27.2} & 65.20 & \underline{58.67} \\
Ours-iter3 & \underline{70.86} & \textbf{69.36} & \underline{61.21} & \textbf{27.3} & \textbf{65.35} & \textbf{58.82} \\
\bottomrule
\end{tabular}
\vspace{-2ex}
\end{table}

\begin{table}
\footnotesize
\setlength{\tabcolsep}{1pt}
\centering
\caption{Results on MiMo-VL-7B-SFT-2508 across 3 iterations.}
\vspace{-2ex}
\label{tab:mimo}
\begin{tabular}{lcccccc}
\toprule
\textbf{Method} & \textbf{MMStar} & \textbf{MMVet} & \textbf{HalluBench} & \textbf{VisuLogic} & \textbf{BLINK} & \textbf{Overall} \\
\midrule
MiMo-VL-7B-SFT-2508  & 72.87 & 66.67 & 60.93 & 20.3 & 62.34 & 56.62 \\
\midrule
Ours-iter1      & 73.07 & \underline{68.53} & \underline{61.28} & 22.1 & \underline{62.65} & 57.53 \\
Ours-iter2      & \underline{73.40} & 67.52 & 61.09 & \underline{23.4} & \textbf{62.96} & \underline{57.68} \\
Ours-iter3      & \textbf{73.47} & \textbf{69.63} & \textbf{61.89} & \textbf{23.7} & 62.39 & \textbf{58.22} \\
\bottomrule
\end{tabular}
\vspace{-4ex}
\end{table}

\begin{figure*}[t!] 
\centering
  \includegraphics[width=1.0\textwidth]{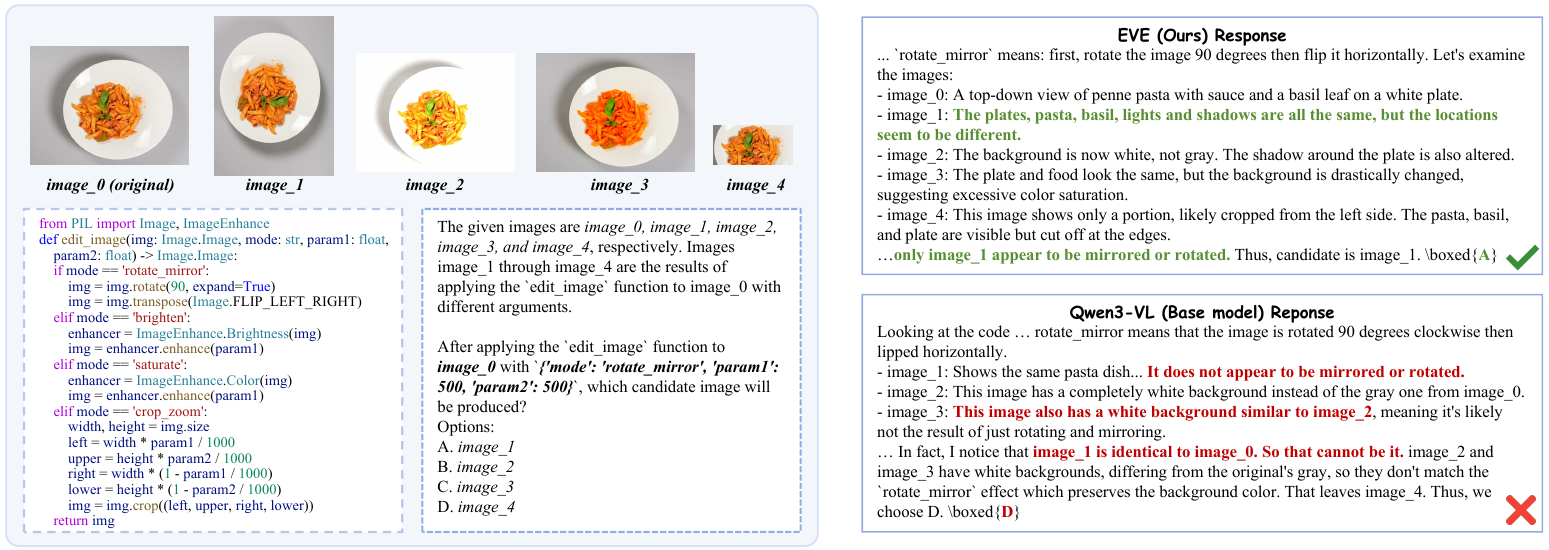}
  \vspace{-4ex}
  \caption{A qualitative example of the Parameter-to-Image task. By generating executable scripts, the Challenger tries visual operations beyond the seed examples (such as color enhancement and mirroring) and combines them to create complex visual transformations. While the baseline model suffers from visual hallucinations, our evolved Solver accurately grounds the code logic in precise visual evidence to make the correct deduction.}
  \label{fig:case}
  \vspace{-2ex}
\end{figure*}

\begin{figure}[t!]
    \centering
    \includegraphics[width=0.75\linewidth]{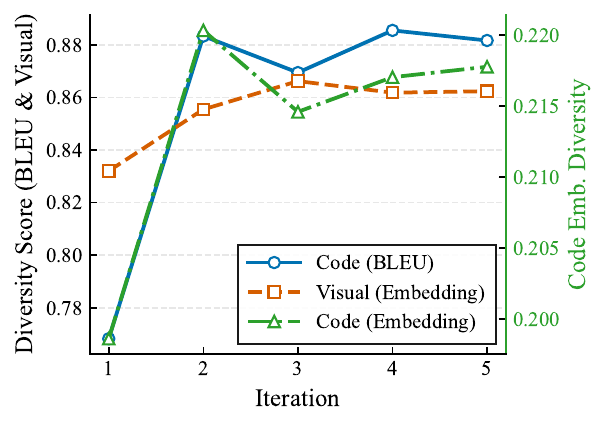}
    \vspace{-2ex}
    \caption{Diversity evolution across iterations. Left axis: code BLEU diversity and visual embedding diversity share the same scale. Right axis: code embedding diversity.}
    \label{fig:diversity_evolution}
    \vspace{-2ex}
\end{figure}

\paragraph{Ablation on challenger reward components.}
We ablate the diversity reward $r_{\text{div}}$ and difficulty reward $r_{\text{diff}}$ to assess their contributions, as shown in Table~\ref{tab:ablation}.
Removing $r_{\text{div}}$ leads to a consistent performance drop across all iterations, with the overall score at iteration 3 decreasing from 70.48 to 69.40 ($-$1.08). This degradation is particularly pronounced on BLINK, where the score drops from 66.81 to 64.80 ($-$2.01), indicating that without diversity regularization, the Challenger generates repetitive transformations that fail to adequately challenge the Solver's capabilities.
Further removing $r_{\text{diff}}$ exacerbates the decline, with the overall score at iteration 3 falling to 68.57 ($-$1.91 compared to the full method). The cumulative effect is evident across all benchmarks: MMStar drops by 1.80 (74.53$\rightarrow$72.73), HallusionBench by 1.75 (62.78$\rightarrow$61.03), and BLINK by 3.01 (66.81$\rightarrow$63.80). Without difficulty calibration, the Challenger produces questions that are either trivially easy or excessively hard, both of which provide weak training signals for the Solver.
These results confirm that both diversity and difficulty rewards are essential for maintaining a rich, appropriately challenging question distribution that drives effective self-evolution.
We do not ablate $r_{\text{format}}$ and $r_{\text{valid}}$, as removing them would prevent the Challenger from generating valid executable code, making it impossible to construct questions for Solver training.

\paragraph{Ablation on Priority queue size $M$.}

We ablate the priority queue size $M$ to examine the effect of retaining high-quality code examples as few-shot context across iterations, as shown in Figure~\ref{fig:ablation_queue}.
Performance increases monotonically with $M$, confirming that a larger pool of high-reward historical programs provides richer few-shot context and enables the Challenger to generate more diverse transformations over iterations.

\paragraph{Ablation on Different Model Scales.}
To verify the generalizability of our framework, we apply it to the smaller Qwen3-VL-4B model, with results shown in Table~\ref{tab:4b_results}.
Our method consistently improves the 4B model across iterations, raising the overall score from 57.54 to 58.82 (+1.28) after 3 iterations.
The smaller gain compared to the 8B model likely reflects the greater difficulty smaller models face in generating sufficiently diverse and high-quality code.

\paragraph{Ablation on Different Model Families.}

We instantiate EVE on MiMo-VL-7B-SFT-2508~\cite{coreteam2025mimovltechnicalreport} to validate the generalizability of our approach across different model families. As shown in Table~\ref{tab:mimo}, EVE consistently improves over the base model across iterations, achieving an overall score of 58.22 at iteration 3, representing a +1.60 gain over MiMo-VL-7B-SFT-2508 (56.62). Gains are observed across most benchmarks, with notable improvements on MMVet (+2.96) and HallusionBench (+0.96), demonstrating that EVE generalizes beyond a single model family.

\subsection{Analysis}
\paragraph{Code Diversity Evolution.}
We track two code-level metrics across iterations (Figure~\ref{fig:diversity_evolution}): (1) \textit{code BLEU diversity}, defined as the mean pairwise BLEU distance between code strings, and (2) \textit{code embedding diversity}, defined as the mean pairwise cosine distance between embedding features (encoded by Qwen3-VL-Embedding-2B~\cite{qwen3vlembedding}) of code strings. \textbf{Both metrics increase consistently across iterations}, confirming that the Challenger generates increasingly varied programs over time. The co-movement of the two metrics further validates that \textbf{BLEU serves as a reliable and computationally efficient proxy for semantic code diversity}.
\paragraph{Visual Diversity Evolution.} As shown in Figure~\ref{fig:diversity_evolution},
We also measure \textit{visual diversity} as the mean pairwise cosine distance between Qwen3-VL-Embedding-2B features of edited images. \textbf{This metric rises in parallel with the code-level diversity metrics}, demonstrating that \textbf{code-level variation directly drives perceptual variation in generated images}. The consistent growth confirms that EVE's self-evolution produces not only structurally diverse programs but also visually distinct training samples.
\paragraph{Qualitative Analysis.}
Crucially, the observed diversity stems not merely from varying parameters within a fixed transformation type (e.g., adjusting rotation angles within the same edit image function, as in the Challenger's prompt template shown in Figure~\ref{fig:teaser}), but from \textbf{the Challenger actively discovering and composing entirely new operation types}. As illustrated in Figure~\ref{fig:case}, the Challenger autonomously invents compound transformations—such as combining rotation with mirroring—that \textbf{go beyond the 4 seed code examples provided at initialization}. These results validate the BLEU-based diversity reward as a semantically grounded signal for promoting cross-modal diversity during RL training.
\vspace{-2ex}

\section{Conclusion}
\label{sec:conclusion}

We presented EVE, a self-evolution framework for multimodal large language models that leverages executable visual transformations as a programmatic external environment. By grounding all supervision in deterministic code execution, EVE eliminates the pseudo-label degradation that plagues confidence-based methods. Meanwhile, the Challenger-Solver dual-policy architecture with multi-dimensional reward-driven co-evolution enables an expandable, adaptive question space, overcoming the diversity and difficulty constraints of template-based approaches. Comprehensive experiments demonstrate consistent improvements across 8 benchmarks, with particularly strong gains on MMVet, VisuLogic and BLINK.





\bibliographystyle{ACM-Reference-Format}
\bibliography{main}


\begin{thebibliography}{50}


\ifx \showCODEN    \undefined \def \showCODEN     #1{\unskip}     \fi
\ifx \showISBNx    \undefined \def \showISBNx     #1{\unskip}     \fi
\ifx \showISBNxiii \undefined \def \showISBNxiii  #1{\unskip}     \fi
\ifx \showISSN     \undefined \def \showISSN      #1{\unskip}     \fi
\ifx \showLCCN     \undefined \def \showLCCN      #1{\unskip}     \fi
\ifx \shownote     \undefined \def \shownote      #1{#1}          \fi
\ifx \showarticletitle \undefined \def \showarticletitle #1{#1}   \fi
\ifx \showURL      \undefined \def \showURL       {\relax}        \fi
\providecommand\bibfield[2]{#2}
\providecommand\bibinfo[2]{#2}
\providecommand\natexlab[1]{#1}
\providecommand\showeprint[2][]{arXiv:#2}

\bibitem[Bae et~al\mbox{.}(2026)]%
        {bae2025online}
\bibfield{author}{\bibinfo{person}{Sanghwan Bae}, \bibinfo{person}{Jiwoo Hong}, \bibinfo{person}{Min~Young Lee}, \bibinfo{person}{Hanbyul Kim}, \bibinfo{person}{JeongYeon Nam}, {and} \bibinfo{person}{Donghyun Kwak}.} \bibinfo{year}{2026}\natexlab{}.
\newblock \showarticletitle{Online difficulty filtering for reasoning oriented reinforcement learning}. In \bibinfo{booktitle}{\emph{Proceedings of the 19th Conference of the European Chapter of the Association for Computational Linguistics (Volume 1: Long Papers)}}. \bibinfo{pages}{700--719}.
\newblock


\bibitem[Bai et~al\mbox{.}(2025)]%
        {bai2025qwen3}
\bibfield{author}{\bibinfo{person}{Shuai Bai}, \bibinfo{person}{Yuxuan Cai}, \bibinfo{person}{Ruizhe Chen}, \bibinfo{person}{Keqin Chen}, \bibinfo{person}{Xionghui Chen}, \bibinfo{person}{Zesen Cheng}, \bibinfo{person}{Lianghao Deng}, \bibinfo{person}{Wei Ding}, \bibinfo{person}{Chang Gao}, \bibinfo{person}{Chunjiang Ge}, {et~al\mbox{.}}} \bibinfo{year}{2025}\natexlab{}.
\newblock \showarticletitle{Qwen3-vl technical report}.
\newblock \bibinfo{journal}{\emph{arXiv preprint arXiv:2511.21631}} (\bibinfo{year}{2025}).
\newblock


\bibitem[Chen et~al\mbox{.}(2024)]%
        {chen2024we}
\bibfield{author}{\bibinfo{person}{Lin Chen}, \bibinfo{person}{Jinsong Li}, \bibinfo{person}{Xiaoyi Dong}, \bibinfo{person}{Pan Zhang}, \bibinfo{person}{Yuhang Zang}, \bibinfo{person}{Zehui Chen}, \bibinfo{person}{Haodong Duan}, \bibinfo{person}{Jiaqi Wang}, \bibinfo{person}{Yu Qiao}, \bibinfo{person}{Dahua Lin}, {et~al\mbox{.}}} \bibinfo{year}{2024}\natexlab{}.
\newblock \showarticletitle{Are we on the right way for evaluating large vision-language models?}
\newblock \bibinfo{journal}{\emph{Advances in Neural Information Processing Systems}}  \bibinfo{volume}{37} (\bibinfo{year}{2024}), \bibinfo{pages}{27056--27087}.
\newblock


\bibitem[Chen et~al\mbox{.}(2025)]%
        {chen2025learning}
\bibfield{author}{\bibinfo{person}{Yang Chen}, \bibinfo{person}{Yufan Shen}, \bibinfo{person}{Wenxuan Huang}, \bibinfo{person}{Sheng Zhou}, \bibinfo{person}{Qunshu Lin}, \bibinfo{person}{Xinyu Cai}, \bibinfo{person}{Zhi Yu}, \bibinfo{person}{Jiajun Bu}, \bibinfo{person}{Botian Shi}, {and} \bibinfo{person}{Yu Qiao}.} \bibinfo{year}{2025}\natexlab{}.
\newblock \showarticletitle{Learning only with images: Visual reinforcement learning with reasoning, rendering, and visual feedback}.
\newblock \bibinfo{journal}{\emph{arXiv preprint arXiv:2507.20766}} (\bibinfo{year}{2025}).
\newblock


\bibitem[Deng et~al\mbox{.}(2025)]%
        {deng2025self}
\bibfield{author}{\bibinfo{person}{Shijian Deng}, \bibinfo{person}{Kai Wang}, \bibinfo{person}{Tianyu Yang}, \bibinfo{person}{Harsh Singh}, {and} \bibinfo{person}{Yapeng Tian}.} \bibinfo{year}{2025}\natexlab{}.
\newblock \showarticletitle{Self-Improvement in Multimodal Large Language Models: A Survey}. In \bibinfo{booktitle}{\emph{Findings of the Association for Computational Linguistics: EMNLP 2025}}. \bibinfo{pages}{1987--2006}.
\newblock


\bibitem[Duan et~al\mbox{.}(2024)]%
        {duan2024vlmevalkit}
\bibfield{author}{\bibinfo{person}{Haodong Duan}, \bibinfo{person}{Junming Yang}, \bibinfo{person}{Yuxuan Qiao}, \bibinfo{person}{Xinyu Fang}, \bibinfo{person}{Lin Chen}, \bibinfo{person}{Yuan Liu}, \bibinfo{person}{Xiaoyi Dong}, \bibinfo{person}{Yuhang Zang}, \bibinfo{person}{Pan Zhang}, \bibinfo{person}{Jiaqi Wang}, {et~al\mbox{.}}} \bibinfo{year}{2024}\natexlab{}.
\newblock \showarticletitle{Vlmevalkit: An open-source toolkit for evaluating large multi-modality models}. In \bibinfo{booktitle}{\emph{Proceedings of the 32nd ACM International Conference on Multimedia}}. \bibinfo{pages}{11198--11201}.
\newblock


\bibitem[Fu et~al\mbox{.}(2024)]%
        {fu2024blink}
\bibfield{author}{\bibinfo{person}{Xingyu Fu}, \bibinfo{person}{Yushi Hu}, \bibinfo{person}{Bangzheng Li}, \bibinfo{person}{Yu Feng}, \bibinfo{person}{Haoyu Wang}, \bibinfo{person}{Xudong Lin}, \bibinfo{person}{Dan Roth}, \bibinfo{person}{Noah~A Smith}, \bibinfo{person}{Wei-Chiu Ma}, {and} \bibinfo{person}{Ranjay Krishna}.} \bibinfo{year}{2024}\natexlab{}.
\newblock \showarticletitle{Blink: Multimodal large language models can see but not perceive}. In \bibinfo{booktitle}{\emph{European Conference on Computer Vision}}. Springer, \bibinfo{pages}{148--166}.
\newblock


\bibitem[Guan et~al\mbox{.}(2024)]%
        {guan2024hallusionbench}
\bibfield{author}{\bibinfo{person}{Tianrui Guan}, \bibinfo{person}{Fuxiao Liu}, \bibinfo{person}{Xiyang Wu}, \bibinfo{person}{Ruiqi Xian}, \bibinfo{person}{Zongxia Li}, \bibinfo{person}{Xiaoyu Liu}, \bibinfo{person}{Xijun Wang}, \bibinfo{person}{Lichang Chen}, \bibinfo{person}{Furong Huang}, \bibinfo{person}{Yaser Yacoob}, {et~al\mbox{.}}} \bibinfo{year}{2024}\natexlab{}.
\newblock \showarticletitle{Hallusionbench: an advanced diagnostic suite for entangled language hallucination and visual illusion in large vision-language models}. In \bibinfo{booktitle}{\emph{Proceedings of the IEEE/CVF conference on computer vision and pattern recognition}}. \bibinfo{pages}{14375--14385}.
\newblock


\bibitem[Guan et~al\mbox{.}(2026)]%
        {guan2026codepercept}
\bibfield{author}{\bibinfo{person}{Tongkun Guan}, \bibinfo{person}{Zhibo Yang}, \bibinfo{person}{Jianqiang Wan}, \bibinfo{person}{Mingkun Yang}, \bibinfo{person}{Zhengtao Guo}, \bibinfo{person}{Zijian Hu}, \bibinfo{person}{Ruilin Luo}, \bibinfo{person}{Ruize Chen}, \bibinfo{person}{Songtao Jiang}, \bibinfo{person}{Peng Wang}, {et~al\mbox{.}}} \bibinfo{year}{2026}\natexlab{}.
\newblock \showarticletitle{CodePercept: Code-Grounded Visual STEM Perception for MLLMs}.
\newblock \bibinfo{journal}{\emph{arXiv preprint arXiv:2603.10757}} (\bibinfo{year}{2026}).
\newblock


\bibitem[He et~al\mbox{.}(2026)]%
        {he2026far}
\bibfield{author}{\bibinfo{person}{Bingxiang He}, \bibinfo{person}{Yuxin Zuo}, \bibinfo{person}{Zeyuan Liu}, \bibinfo{person}{Shangziqi Zhao}, \bibinfo{person}{Zixuan Fu}, \bibinfo{person}{Junlin Yang}, \bibinfo{person}{Cheng Qian}, \bibinfo{person}{Kaiyan Zhang}, \bibinfo{person}{Yuchen Fan}, \bibinfo{person}{Ganqu Cui}, {et~al\mbox{.}}} \bibinfo{year}{2026}\natexlab{}.
\newblock \showarticletitle{How Far Can Unsupervised RLVR Scale LLM Training?}
\newblock \bibinfo{journal}{\emph{arXiv preprint arXiv:2603.08660}} (\bibinfo{year}{2026}).
\newblock


\bibitem[He et~al\mbox{.}(2025)]%
        {he2025visplay}
\bibfield{author}{\bibinfo{person}{Yicheng He}, \bibinfo{person}{Chengsong Huang}, \bibinfo{person}{Zongxia Li}, \bibinfo{person}{Jiaxin Huang}, {and} \bibinfo{person}{Yonghui Yang}.} \bibinfo{year}{2025}\natexlab{}.
\newblock \showarticletitle{Visplay: Self-evolving vision-language models from images}.
\newblock \bibinfo{journal}{\emph{arXiv preprint arXiv:2511.15661}} (\bibinfo{year}{2025}).
\newblock


\bibitem[Hong et~al\mbox{.}(2025)]%
        {hong2025deepeyesv2}
\bibfield{author}{\bibinfo{person}{Jack Hong}, \bibinfo{person}{Chenxiao Zhao}, \bibinfo{person}{ChengLin Zhu}, \bibinfo{person}{Weiheng Lu}, \bibinfo{person}{Guohai Xu}, {and} \bibinfo{person}{Xing Yu}.} \bibinfo{year}{2025}\natexlab{}.
\newblock \showarticletitle{DeepEyesV2: Toward Agentic Multimodal Model}.
\newblock \bibinfo{journal}{\emph{arXiv preprint arXiv:2511.05271}} (\bibinfo{year}{2025}).
\newblock


\bibitem[Huang et~al\mbox{.}(2025)]%
        {huang2025rzero}
\bibfield{author}{\bibinfo{person}{Chengsong Huang}, \bibinfo{person}{Lantao Yu}, \bibinfo{person}{Yicheng He}, \bibinfo{person}{Zongxia Li}, \bibinfo{person}{Jiaxin Huang}, {and} \bibinfo{person}{Yonghui Yang}.} \bibinfo{year}{2025}\natexlab{}.
\newblock \showarticletitle{R-Zero: Self-Evolving Reasoning LLM from Zero Data}.
\newblock \bibinfo{journal}{\emph{arXiv preprint arXiv:2508.05004}} (\bibinfo{year}{2025}).
\newblock


\bibitem[Hurst et~al\mbox{.}(2024)]%
        {hurst2024gpt}
\bibfield{author}{\bibinfo{person}{Aaron Hurst}, \bibinfo{person}{Adam Lerer}, \bibinfo{person}{Adam~P Goucher}, \bibinfo{person}{Adam Perelman}, \bibinfo{person}{Aditya Ramesh}, \bibinfo{person}{Aidan Clark}, \bibinfo{person}{AJ Ostrow}, \bibinfo{person}{Akila Welihinda}, \bibinfo{person}{Alan Hayes}, \bibinfo{person}{Alec Radford}, {et~al\mbox{.}}} \bibinfo{year}{2024}\natexlab{}.
\newblock \showarticletitle{Gpt-4o system card}.
\newblock \bibinfo{journal}{\emph{arXiv preprint arXiv:2410.21276}} (\bibinfo{year}{2024}).
\newblock


\bibitem[Jeddi et~al\mbox{.}(2025)]%
        {jeddi2025puzzle}
\bibfield{author}{\bibinfo{person}{Ahmadreza Jeddi}, \bibinfo{person}{Hakki~Can Karaimer}, \bibinfo{person}{Hue Nguyen}, \bibinfo{person}{Zhongling Wang}, \bibinfo{person}{Ke Zhao}, \bibinfo{person}{Javad Rajabi}, \bibinfo{person}{Ran Zhang}, \bibinfo{person}{Raghav Goyal}, \bibinfo{person}{Babak Taati}, {and} \bibinfo{person}{Radek Grzeszczuk}.} \bibinfo{year}{2025}\natexlab{}.
\newblock \showarticletitle{Puzzle Curriculum GRPO for Vision-Centric Reasoning}.
\newblock \bibinfo{journal}{\emph{arXiv preprint arXiv:2512.14944}} (\bibinfo{year}{2025}).
\newblock


\bibitem[Khan et~al\mbox{.}(2024)]%
        {khan2024dataenvgym}
\bibfield{author}{\bibinfo{person}{Zaid Khan}, \bibinfo{person}{Elias Stengel-Eskin}, \bibinfo{person}{Jaemin Cho}, {and} \bibinfo{person}{Mohit Bansal}.} \bibinfo{year}{2024}\natexlab{}.
\newblock \showarticletitle{Dataenvgym: Data generation agents in teacher environments with student feedback}.
\newblock \bibinfo{journal}{\emph{arXiv preprint arXiv:2410.06215}} (\bibinfo{year}{2024}).
\newblock


\bibitem[Kwon et~al\mbox{.}(2023)]%
        {kwon2023efficient}
\bibfield{author}{\bibinfo{person}{Woosuk Kwon}, \bibinfo{person}{Zhuohan Li}, \bibinfo{person}{Siyuan Zhuang}, \bibinfo{person}{Ying Sheng}, \bibinfo{person}{Lianmin Zheng}, \bibinfo{person}{Cody~Hao Yu}, \bibinfo{person}{Joseph Gonzalez}, \bibinfo{person}{Hao Zhang}, {and} \bibinfo{person}{Ion Stoica}.} \bibinfo{year}{2023}\natexlab{}.
\newblock \showarticletitle{Efficient memory management for large language model serving with pagedattention}. In \bibinfo{booktitle}{\emph{Proceedings of the 29th symposium on operating systems principles}}. \bibinfo{pages}{611--626}.
\newblock


\bibitem[Li et~al\mbox{.}(2024)]%
        {li2024llava}
\bibfield{author}{\bibinfo{person}{Bo Li}, \bibinfo{person}{Yuanhan Zhang}, \bibinfo{person}{Dong Guo}, \bibinfo{person}{Renrui Zhang}, \bibinfo{person}{Feng Li}, \bibinfo{person}{Hao Zhang}, \bibinfo{person}{Kaichen Zhang}, \bibinfo{person}{Peiyuan Zhang}, \bibinfo{person}{Yanwei Li}, \bibinfo{person}{Ziwei Liu}, {et~al\mbox{.}}} \bibinfo{year}{2024}\natexlab{}.
\newblock \showarticletitle{Llava-onevision: Easy visual task transfer}.
\newblock \bibinfo{journal}{\emph{arXiv preprint arXiv:2408.03326}} (\bibinfo{year}{2024}).
\newblock


\bibitem[Li et~al\mbox{.}(2026b)]%
        {qwen3vlembedding}
\bibfield{author}{\bibinfo{person}{Mingxin Li}, \bibinfo{person}{Yanzhao Zhang}, \bibinfo{person}{Dingkun Long}, \bibinfo{person}{Chen Keqin}, \bibinfo{person}{Sibo Song}, \bibinfo{person}{Shuai Bai}, \bibinfo{person}{Zhibo Yang}, \bibinfo{person}{Pengjun Xie}, \bibinfo{person}{An Yang}, \bibinfo{person}{Dayiheng Liu}, \bibinfo{person}{Jingren Zhou}, {and} \bibinfo{person}{Junyang Lin}.} \bibinfo{year}{2026}\natexlab{b}.
\newblock \showarticletitle{Qwen3-VL-Embedding and Qwen3-VL-Reranker: A Unified Framework for State-of-the-Art Multimodal Retrieval and Ranking}.
\newblock \bibinfo{journal}{\emph{arXiv preprint arXiv:2601.04720}} (\bibinfo{year}{2026}).
\newblock


\bibitem[Li et~al\mbox{.}(2026a)]%
        {li2026mm}
\bibfield{author}{\bibinfo{person}{Zongxia Li}, \bibinfo{person}{Hongyang Du}, \bibinfo{person}{Chengsong Huang}, \bibinfo{person}{Xiyang Wu}, \bibinfo{person}{Lantao Yu}, \bibinfo{person}{Yicheng He}, \bibinfo{person}{Jing Xie}, \bibinfo{person}{Xiaomin Wu}, \bibinfo{person}{Zhichao Liu}, \bibinfo{person}{Jiarui Zhang}, {et~al\mbox{.}}} \bibinfo{year}{2026}\natexlab{a}.
\newblock \showarticletitle{MM-Zero: Self-Evolving Multi-Model Vision Language Models From Zero Data}.
\newblock \bibinfo{journal}{\emph{arXiv preprint arXiv:2603.09206}} (\bibinfo{year}{2026}).
\newblock


\bibitem[Li et~al\mbox{.}(2025)]%
        {li2025selfrewardingvisionlanguagemodelreasoning}
\bibfield{author}{\bibinfo{person}{Zongxia Li}, \bibinfo{person}{Wenhao Yu}, \bibinfo{person}{Chengsong Huang}, \bibinfo{person}{Rui Liu}, \bibinfo{person}{Zhenwen Liang}, \bibinfo{person}{Fuxiao Liu}, \bibinfo{person}{Jingxi Che}, \bibinfo{person}{Dian Yu}, \bibinfo{person}{Jordan Boyd-Graber}, \bibinfo{person}{Haitao Mi}, {and} \bibinfo{person}{Dong Yu}.} \bibinfo{year}{2025}\natexlab{}.
\newblock \bibinfo{title}{Self-Rewarding Vision-Language Model via Reasoning Decomposition}.
\newblock
\showeprint[arxiv]{2508.19652}~[cs.CV]
\urldef\tempurl%
\url{https://arxiv.org/abs/2508.19652}
\showURL{%
\tempurl}


\bibitem[Liang et~al\mbox{.}(2023)]%
        {liang2023code}
\bibfield{author}{\bibinfo{person}{Jacky Liang}, \bibinfo{person}{Wenlong Huang}, \bibinfo{person}{Fei Xia}, \bibinfo{person}{Peng Xu}, \bibinfo{person}{Karol Hausman}, \bibinfo{person}{Brian Ichter}, \bibinfo{person}{Pete Florence}, {and} \bibinfo{person}{Andy Zeng}.} \bibinfo{year}{2023}\natexlab{}.
\newblock \showarticletitle{Code as policies: Language model programs for embodied control}. In \bibinfo{booktitle}{\emph{2023 IEEE International conference on robotics and automation (ICRA)}}. IEEE, \bibinfo{pages}{9493--9500}.
\newblock


\bibitem[Liu et~al\mbox{.}(2024)]%
        {liu2024diving}
\bibfield{author}{\bibinfo{person}{Wei Liu}, \bibinfo{person}{Junlong Li}, \bibinfo{person}{Xiwen Zhang}, \bibinfo{person}{Fan Zhou}, \bibinfo{person}{Yu Cheng}, {and} \bibinfo{person}{Junxian He}.} \bibinfo{year}{2024}\natexlab{}.
\newblock \showarticletitle{Diving into self-evolving training for multimodal reasoning}.
\newblock \bibinfo{journal}{\emph{arXiv preprint arXiv:2412.17451}} (\bibinfo{year}{2024}).
\newblock


\bibitem[Liu et~al\mbox{.}(2026)]%
        {liu2026self}
\bibfield{author}{\bibinfo{person}{Wei Liu}, \bibinfo{person}{Siya Qi}, \bibinfo{person}{Yali Du}, {and} \bibinfo{person}{Yulan He}.} \bibinfo{year}{2026}\natexlab{}.
\newblock \showarticletitle{Self-Play Only Evolves When Self-Synthetic Pipeline Ensures Learnable Information Gain}.
\newblock \bibinfo{journal}{\emph{arXiv preprint arXiv:2603.02218}} (\bibinfo{year}{2026}).
\newblock


\bibitem[Liu et~al\mbox{.}(2025)]%
        {liu2025spatial}
\bibfield{author}{\bibinfo{person}{Yuhong Liu}, \bibinfo{person}{Beichen Zhang}, \bibinfo{person}{Yuhang Zang}, \bibinfo{person}{Yuhang Cao}, \bibinfo{person}{Long Xing}, \bibinfo{person}{Xiaoyi Dong}, \bibinfo{person}{Haodong Duan}, \bibinfo{person}{Dahua Lin}, {and} \bibinfo{person}{Jiaqi Wang}.} \bibinfo{year}{2025}\natexlab{}.
\newblock \showarticletitle{Spatial-ssrl: Enhancing spatial understanding via self-supervised reinforcement learning}.
\newblock \bibinfo{journal}{\emph{arXiv preprint arXiv:2510.27606}} (\bibinfo{year}{2025}).
\newblock


\bibitem[Lu et~al\mbox{.}(2023)]%
        {lu2023mathvista}
\bibfield{author}{\bibinfo{person}{Pan Lu}, \bibinfo{person}{Hritik Bansal}, \bibinfo{person}{Tony Xia}, \bibinfo{person}{Jiacheng Liu}, \bibinfo{person}{Chunyuan Li}, \bibinfo{person}{Hannaneh Hajishirzi}, \bibinfo{person}{Hao Cheng}, \bibinfo{person}{Kai-Wei Chang}, \bibinfo{person}{Michel Galley}, {and} \bibinfo{person}{Jianfeng Gao}.} \bibinfo{year}{2023}\natexlab{}.
\newblock \showarticletitle{Mathvista: Evaluating Math Reasoning of foundation models in visual contexts}.
\newblock \bibinfo{journal}{\emph{arXiv preprint arXiv:2310.02255}} (\bibinfo{year}{2023}).
\newblock


\bibitem[Papineni et~al\mbox{.}(2002)]%
        {papineni2002bleu}
\bibfield{author}{\bibinfo{person}{Kishore Papineni}, \bibinfo{person}{Salim Roukos}, \bibinfo{person}{Todd Ward}, {and} \bibinfo{person}{Wei-Jing Zhu}.} \bibinfo{year}{2002}\natexlab{}.
\newblock \showarticletitle{Bleu: a method for automatic evaluation of machine translation}. In \bibinfo{booktitle}{\emph{Proceedings of the 40th annual meeting of the Association for Computational Linguistics}}. \bibinfo{pages}{311--318}.
\newblock


\bibitem[Qian et~al\mbox{.}(2024)]%
        {qian2024mia}
\bibfield{author}{\bibinfo{person}{Yusu Qian}, \bibinfo{person}{Hanrong Ye}, \bibinfo{person}{Jean-Philippe Fauconnier}, \bibinfo{person}{Peter Grasch}, \bibinfo{person}{Yinfei Yang}, {and} \bibinfo{person}{Zhe Gan}.} \bibinfo{year}{2024}\natexlab{}.
\newblock \showarticletitle{Mia-bench: Towards better instruction following evaluation of multimodal llms}.
\newblock \bibinfo{journal}{\emph{arXiv preprint arXiv:2407.01509}} (\bibinfo{year}{2024}).
\newblock


\bibitem[Radford et~al\mbox{.}(2021)]%
        {radford2021learning}
\bibfield{author}{\bibinfo{person}{Alec Radford}, \bibinfo{person}{Jong~Wook Kim}, \bibinfo{person}{Chris Hallacy}, \bibinfo{person}{Aditya Ramesh}, \bibinfo{person}{Gabriel Goh}, \bibinfo{person}{Sandhini Agarwal}, \bibinfo{person}{Girish Sastry}, \bibinfo{person}{Amanda Askell}, \bibinfo{person}{Pamela Mishkin}, \bibinfo{person}{Jack Clark}, {et~al\mbox{.}}} \bibinfo{year}{2021}\natexlab{}.
\newblock \showarticletitle{Learning transferable visual models from natural language supervision}. In \bibinfo{booktitle}{\emph{International conference on machine learning}}. PmLR, \bibinfo{pages}{8748--8763}.
\newblock


\bibitem[Shao et~al\mbox{.}(2024)]%
        {shao2024deepseekmath}
\bibfield{author}{\bibinfo{person}{Zhihong Shao}, \bibinfo{person}{Peiyi Wang}, \bibinfo{person}{Qihao Zhu}, \bibinfo{person}{Runxin Xu}, \bibinfo{person}{Junxiao Song}, \bibinfo{person}{Xiao Bi}, \bibinfo{person}{Haowei Zhang}, \bibinfo{person}{Mingchuan Zhang}, \bibinfo{person}{YK Li}, \bibinfo{person}{Yang Wu}, {et~al\mbox{.}}} \bibinfo{year}{2024}\natexlab{}.
\newblock \showarticletitle{Deepseekmath: Pushing the limits of Math Reasoning in open language models}.
\newblock \bibinfo{journal}{\emph{arXiv preprint arXiv:2402.03300}} (\bibinfo{year}{2024}).
\newblock


\bibitem[Sheng et~al\mbox{.}(2024)]%
        {sheng2024hybridflow}
\bibfield{author}{\bibinfo{person}{Guangming Sheng}, \bibinfo{person}{Chi Zhang}, \bibinfo{person}{Zilingfeng Ye}, \bibinfo{person}{Xibin Wu}, \bibinfo{person}{Wang Zhang}, \bibinfo{person}{Ru Zhang}, \bibinfo{person}{Yanghua Peng}, \bibinfo{person}{Haibin Lin}, {and} \bibinfo{person}{Chuan Wu}.} \bibinfo{year}{2024}\natexlab{}.
\newblock \showarticletitle{HybridFlow: A Flexible and Efficient RLHF Framework}.
\newblock \bibinfo{journal}{\emph{arXiv preprint arXiv: 2409.19256}} (\bibinfo{year}{2024}).
\newblock


\bibitem[Singh et~al\mbox{.}(2025)]%
        {singh2025openai}
\bibfield{author}{\bibinfo{person}{Aaditya Singh}, \bibinfo{person}{Adam Fry}, \bibinfo{person}{Adam Perelman}, \bibinfo{person}{Adam Tart}, \bibinfo{person}{Adi Ganesh}, \bibinfo{person}{Ahmed El-Kishky}, \bibinfo{person}{Aidan McLaughlin}, \bibinfo{person}{Aiden Low}, \bibinfo{person}{AJ Ostrow}, \bibinfo{person}{Akhila Ananthram}, {et~al\mbox{.}}} \bibinfo{year}{2025}\natexlab{}.
\newblock \showarticletitle{Openai gpt-5 system card}.
\newblock \bibinfo{journal}{\emph{arXiv preprint arXiv:2601.03267}} (\bibinfo{year}{2025}).
\newblock


\bibitem[Song et~al\mbox{.}(2025)]%
        {song2025codedance}
\bibfield{author}{\bibinfo{person}{Qi Song}, \bibinfo{person}{Honglin Li}, \bibinfo{person}{Yingchen Yu}, \bibinfo{person}{Haoyi Zhou}, \bibinfo{person}{Lin Yang}, \bibinfo{person}{Song Bai}, \bibinfo{person}{Qi She}, \bibinfo{person}{Zilong Huang}, {and} \bibinfo{person}{Yunqing Zhao}.} \bibinfo{year}{2025}\natexlab{}.
\newblock \showarticletitle{CodeDance: A Dynamic Tool-integrated MLLM for Executable Visual Reasoning}.
\newblock \bibinfo{journal}{\emph{arXiv preprint arXiv:2512.17312}} (\bibinfo{year}{2025}).
\newblock


\bibitem[Sunil et~al\mbox{.}(2026)]%
        {sunil2026ireasoner}
\bibfield{author}{\bibinfo{person}{Meghana Sunil}, \bibinfo{person}{Manikandarajan Venmathimaran}, {and} \bibinfo{person}{Muthu~Subash Kavitha}.} \bibinfo{year}{2026}\natexlab{}.
\newblock \showarticletitle{iReasoner: Trajectory-Aware Intrinsic Reasoning Supervision for Self-Evolving Large Multimodal Models}.
\newblock \bibinfo{journal}{\emph{arXiv preprint arXiv:2601.05877}} (\bibinfo{year}{2026}).
\newblock


\bibitem[Sur{\'\i}s et~al\mbox{.}(2023)]%
        {suris2023vipergpt}
\bibfield{author}{\bibinfo{person}{D{\'\i}dac Sur{\'\i}s}, \bibinfo{person}{Sachit Menon}, {and} \bibinfo{person}{Carl Vondrick}.} \bibinfo{year}{2023}\natexlab{}.
\newblock \showarticletitle{Vipergpt: Visual inference via python execution for reasoning}. In \bibinfo{booktitle}{\emph{Proceedings of the IEEE/CVF international conference on computer vision}}. \bibinfo{pages}{11888--11898}.
\newblock


\bibitem[Thawakar et~al\mbox{.}(2025)]%
        {thawakar2025evolmm}
\bibfield{author}{\bibinfo{person}{Omkar Thawakar}, \bibinfo{person}{Shravan Venkatraman}, \bibinfo{person}{Ritesh Thawkar}, \bibinfo{person}{Abdelrahman Shaker}, \bibinfo{person}{Hisham Cholakkal}, \bibinfo{person}{Rao~Muhammad Anwer}, \bibinfo{person}{Salman Khan}, {and} \bibinfo{person}{Fahad Khan}.} \bibinfo{year}{2025}\natexlab{}.
\newblock \showarticletitle{Evolmm: Self-evolving large multimodal models with continuous rewards}.
\newblock \bibinfo{journal}{\emph{arXiv preprint arXiv:2511.16672}} (\bibinfo{year}{2025}).
\newblock


\bibitem[Wang et~al\mbox{.}(2024)]%
        {wang2024muirbench}
\bibfield{author}{\bibinfo{person}{Fei Wang}, \bibinfo{person}{Xingyu Fu}, \bibinfo{person}{James~Y Huang}, \bibinfo{person}{Zekun Li}, \bibinfo{person}{Qin Liu}, \bibinfo{person}{Xiaogeng Liu}, \bibinfo{person}{Mingyu~Derek Ma}, \bibinfo{person}{Nan Xu}, \bibinfo{person}{Wenxuan Zhou}, \bibinfo{person}{Kai Zhang}, {et~al\mbox{.}}} \bibinfo{year}{2024}\natexlab{}.
\newblock \showarticletitle{Muirbench: A comprehensive benchmark for robust multi-image understanding}.
\newblock \bibinfo{journal}{\emph{arXiv preprint arXiv:2406.09411}} (\bibinfo{year}{2024}).
\newblock


\bibitem[Wang et~al\mbox{.}(2026)]%
        {wang2026v}
\bibfield{author}{\bibinfo{person}{Han Wang}, \bibinfo{person}{Yi Yang}, \bibinfo{person}{Jingyuan Hu}, \bibinfo{person}{Minfeng Zhu}, {and} \bibinfo{person}{Wei Chen}.} \bibinfo{year}{2026}\natexlab{}.
\newblock \showarticletitle{V-Zero: Self-Improving Multimodal Reasoning with Zero Annotation}.
\newblock \bibinfo{journal}{\emph{arXiv preprint arXiv:2601.10094}} (\bibinfo{year}{2026}).
\newblock


\bibitem[Wang et~al\mbox{.}(2025)]%
        {wang2025jigsawr1}
\bibfield{author}{\bibinfo{person}{Zifu Wang}, \bibinfo{person}{Junyi Zhu}, \bibinfo{person}{Bo Tang}, \bibinfo{person}{Zhiyu Li}, \bibinfo{person}{Feiyu Xiong}, \bibinfo{person}{Jiaqian Yu}, {and} \bibinfo{person}{Matthew~B. Blaschko}.} \bibinfo{year}{2025}\natexlab{}.
\newblock \bibinfo{title}{Jigsaw-R1: A Study of Rule-based Visual Reinforcement Learning with Jigsaw Puzzles}.
\newblock


\bibitem[Xiaomi(2025)]%
        {coreteam2025mimovltechnicalreport}
\bibfield{author}{\bibinfo{person}{LLM-Core-Team Xiaomi}.} \bibinfo{year}{2025}\natexlab{}.
\newblock \bibinfo{title}{MiMo-VL Technical Report}.
\newblock
\showeprint[arxiv]{2506.03569}~[cs.CL]
\urldef\tempurl%
\url{https://arxiv.org/abs/2506.03569}
\showURL{%
\tempurl}


\bibitem[Xu et~al\mbox{.}(2025)]%
        {xu2025visulogic}
\bibfield{author}{\bibinfo{person}{Weiye Xu}, \bibinfo{person}{Jiahao Wang}, \bibinfo{person}{Weiyun Wang}, \bibinfo{person}{Zhe Chen}, \bibinfo{person}{Wengang Zhou}, \bibinfo{person}{Aijun Yang}, \bibinfo{person}{Lewei Lu}, \bibinfo{person}{Houqiang Li}, \bibinfo{person}{Xiaohua Wang}, \bibinfo{person}{Xizhou Zhu}, {et~al\mbox{.}}} \bibinfo{year}{2025}\natexlab{}.
\newblock \showarticletitle{Visulogic: A benchmark for evaluating visual reasoning in multi-modal large language models}.
\newblock \bibinfo{journal}{\emph{arXiv preprint arXiv:2504.15279}} (\bibinfo{year}{2025}).
\newblock


\bibitem[Yu et~al\mbox{.}(2025)]%
        {yu2025dapo}
\bibfield{author}{\bibinfo{person}{Qiying Yu}, \bibinfo{person}{Zheng Zhang}, \bibinfo{person}{Ruofei Zhu}, \bibinfo{person}{Yufeng Yuan}, \bibinfo{person}{Xiaochen Zuo}, \bibinfo{person}{Yu Yue}, \bibinfo{person}{Weinan Dai}, \bibinfo{person}{Tiantian Fan}, \bibinfo{person}{Gaohong Liu}, \bibinfo{person}{Lingjun Liu}, {et~al\mbox{.}}} \bibinfo{year}{2025}\natexlab{}.
\newblock \showarticletitle{Dapo: An open-source llm reinforcement learning system at scale}.
\newblock \bibinfo{journal}{\emph{arXiv preprint arXiv:2503.14476}} (\bibinfo{year}{2025}).
\newblock


\bibitem[Yu et~al\mbox{.}(2023)]%
        {yu2023mm}
\bibfield{author}{\bibinfo{person}{Weihao Yu}, \bibinfo{person}{Zhengyuan Yang}, \bibinfo{person}{Linjie Li}, \bibinfo{person}{Jianfeng Wang}, \bibinfo{person}{Kevin Lin}, \bibinfo{person}{Zicheng Liu}, \bibinfo{person}{Xinchao Wang}, {and} \bibinfo{person}{Lijuan Wang}.} \bibinfo{year}{2023}\natexlab{}.
\newblock \showarticletitle{Mm-vet: Evaluating large multimodal models for integrated capabilities}.
\newblock \bibinfo{journal}{\emph{arXiv preprint arXiv:2308.02490}} (\bibinfo{year}{2023}).
\newblock


\bibitem[Zeng et~al\mbox{.}(2025)]%
        {zeng2025agentic}
\bibfield{author}{\bibinfo{person}{Yu Zeng}, \bibinfo{person}{Wenxuan Huang}, \bibinfo{person}{Shiting Huang}, \bibinfo{person}{Xikun Bao}, \bibinfo{person}{Yukun Qi}, \bibinfo{person}{Yiming Zhao}, \bibinfo{person}{Qiuchen Wang}, \bibinfo{person}{Lin Chen}, \bibinfo{person}{Zehui Chen}, \bibinfo{person}{Huaian Chen}, {et~al\mbox{.}}} \bibinfo{year}{2025}\natexlab{}.
\newblock \showarticletitle{Agentic Jigsaw Interaction Learning for Enhancing Visual Perception and Reasoning in Vision-Language Models}.
\newblock \bibinfo{journal}{\emph{arXiv preprint arXiv:2510.01304}} (\bibinfo{year}{2025}).
\newblock


\bibitem[Zhang et~al\mbox{.}(2025a)]%
        {zhang2025viper}
\bibfield{author}{\bibinfo{person}{Juntian Zhang}, \bibinfo{person}{Song Jin}, \bibinfo{person}{Chuanqi Cheng}, \bibinfo{person}{Yuhan Liu}, \bibinfo{person}{Yankai Lin}, \bibinfo{person}{Xun Zhang}, \bibinfo{person}{Yufei Zhang}, \bibinfo{person}{Fei Jiang}, \bibinfo{person}{Guojun Yin}, \bibinfo{person}{Wei Lin}, {et~al\mbox{.}}} \bibinfo{year}{2025}\natexlab{a}.
\newblock \showarticletitle{Viper: Empowering the self-evolution of visual perception abilities in vision-language model}.
\newblock \bibinfo{journal}{\emph{arXiv preprint arXiv:2510.24285}} (\bibinfo{year}{2025}).
\newblock


\bibitem[Zhang et~al\mbox{.}(2025b)]%
        {zhang2025thyme}
\bibfield{author}{\bibinfo{person}{Yi-Fan Zhang}, \bibinfo{person}{Xingyu Lu}, \bibinfo{person}{Shukang Yin}, \bibinfo{person}{Chaoyou Fu}, \bibinfo{person}{Wei Chen}, \bibinfo{person}{Xiao Hu}, \bibinfo{person}{Bin Wen}, \bibinfo{person}{Kaiyu Jiang}, \bibinfo{person}{Changyi Liu}, \bibinfo{person}{Tianke Zhang}, {et~al\mbox{.}}} \bibinfo{year}{2025}\natexlab{b}.
\newblock \showarticletitle{Thyme: Think Beyond Images}.
\newblock \bibinfo{journal}{\emph{arXiv preprint arXiv:2508.11630}} (\bibinfo{year}{2025}).
\newblock


\bibitem[Zhao et~al\mbox{.}(2025a)]%
        {zhao2025absolute}
\bibfield{author}{\bibinfo{person}{Andrew Zhao}, \bibinfo{person}{Yiran Wu}, \bibinfo{person}{Yang Yue}, \bibinfo{person}{Tong Wu}, \bibinfo{person}{Quentin Xu}, \bibinfo{person}{Matthieu Lin}, \bibinfo{person}{Shenzhi Wang}, \bibinfo{person}{Qingyun Wu}, \bibinfo{person}{Zilong Zheng}, {and} \bibinfo{person}{Gao Huang}.} \bibinfo{year}{2025}\natexlab{a}.
\newblock \showarticletitle{Absolute zero: Reinforced self-play reasoning with zero data}.
\newblock \bibinfo{journal}{\emph{arXiv preprint arXiv:2505.03335}} (\bibinfo{year}{2025}).
\newblock


\bibitem[Zhao et~al\mbox{.}(2026)]%
        {zhao2026pyvision}
\bibfield{author}{\bibinfo{person}{Shitian Zhao}, \bibinfo{person}{Shaoheng Lin}, \bibinfo{person}{Ming Li}, \bibinfo{person}{Haoquan Zhang}, \bibinfo{person}{Wenshuo Peng}, \bibinfo{person}{Kaipeng Zhang}, {and} \bibinfo{person}{Chen Wei}.} \bibinfo{year}{2026}\natexlab{}.
\newblock \showarticletitle{PyVision-RL: Forging Open Agentic Vision Models via RL}.
\newblock \bibinfo{journal}{\emph{arXiv preprint arXiv:2602.20739}} (\bibinfo{year}{2026}).
\newblock


\bibitem[Zhao et~al\mbox{.}(2025b)]%
        {zhao2025pyvision}
\bibfield{author}{\bibinfo{person}{Shitian Zhao}, \bibinfo{person}{Haoquan Zhang}, \bibinfo{person}{Shaoheng Lin}, \bibinfo{person}{Ming Li}, \bibinfo{person}{Qilong Wu}, \bibinfo{person}{Kaipeng Zhang}, {and} \bibinfo{person}{Chen Wei}.} \bibinfo{year}{2025}\natexlab{b}.
\newblock \showarticletitle{Pyvision: Agentic vision with dynamic tooling}.
\newblock \bibinfo{journal}{\emph{arXiv preprint arXiv:2507.07998}} (\bibinfo{year}{2025}).
\newblock


\bibitem[Zhu et~al\mbox{.}(2025)]%
        {zhu2025internvl3}
\bibfield{author}{\bibinfo{person}{Jinguo Zhu}, \bibinfo{person}{Weiyun Wang}, \bibinfo{person}{Zhe Chen}, \bibinfo{person}{Zhaoyang Liu}, \bibinfo{person}{Shenglong Ye}, \bibinfo{person}{Lixin Gu}, \bibinfo{person}{Hao Tian}, \bibinfo{person}{Yuchen Duan}, \bibinfo{person}{Weijie Su}, \bibinfo{person}{Jie Shao}, {et~al\mbox{.}}} \bibinfo{year}{2025}\natexlab{}.
\newblock \showarticletitle{Internvl3: Exploring advanced training and test-time recipes for open-source multimodal models}.
\newblock \bibinfo{journal}{\emph{arXiv preprint arXiv:2504.10479}} (\bibinfo{year}{2025}).
\newblock


\end{thebibliography}

\clearpage
\appendix
\appendix

\section{Implementation Details}
\label{sec:appendix_implementation}

\subsection{Sandbox Execution}

Code execution is performed in a restricted Python sandbox, which is adapted from the Thyme~\cite{zhang2025thyme} sandbox environment, with the following constraints:
\begin{itemize}
\item Timeout: 5 seconds per execution
\item Prohibited operations: network access, subprocess calls, random module usage
\item Random seed is fixed to ensure reproducibility
\item Code length limit: 2000 characters for the generated code
\end{itemize}

\subsection{Training Hyperparameters}

The hyperparameters used for training the self-evolution loop are detailed in Table~\ref{tab:hyperparams}.
\label{sec:appendix_hyperparams}
\begin{table}
\centering
\caption{Training hyperparameters for the self-evolution loop.}
\label{tab:hyperparams}
\small
\begin{tabular}{lc}
\toprule
\textbf{Hyperparameter} & \textbf{Value} \\
\midrule
\multicolumn{2}{c}{\textit{General Training}} \\
Total iterations $T$ & 3 \\
Training steps per iteration $N_{\text{step}}$ & 10 \\
Learning rate & 1e-6 \\
Training batch size & 128 \\
\midrule
\multicolumn{2}{c}{\textit{RL Optimization (Eq.~\ref{eq:training_objective})}} \\
KL coefficient $\beta$ & 0.01 \\
KL loss type & low\_var\_kl \\
Clip range $\epsilon_{\text{low}}$ & 0.28 \\
Clip range $\epsilon_{\text{high}}$ & 0.20 \\
\midrule
\multicolumn{2}{c}{\textit{Sampling Parameters}} \\
Rollout batch size $B$ & 128 \\
Temperature & 1.0 \\
Top-p & 1.0 \\
Top-k & 40 \\
Presence penalty & 2.0 \\
Max prompt length & 8192 \\
Max response length & 2048 \\
\midrule
\multicolumn{2}{c}{\textit{Challenger Training}} \\
Solver sampling rounds $K$ & 6 \\
Rollout sampling $G$ & 4 \\
GPUs & 4 \\
Format weight $\lambda_{\text{format}}$ & 0.2 \\
Validity weight $\lambda_{\text{valid}}$ & 0.4 \\
Difficulty weight $\lambda_{\text{diff}}$ & 0.4 \\
Diversity weight $\lambda_{\text{div}}$ & 0.3 \\
\midrule
\multicolumn{2}{c}{\textit{Solver Training}} \\
Rollout sampling $G$ & 8 \\
GPUs & 8 \\
Format weight $\omega_{\text{format}}$ & 0.2 \\
Accuracy weight $\omega_{\text{acc}}$ & 0.8 \\
\midrule
\multicolumn{2}{c}{\textit{Method-related}} \\
BLEU threshold for deduplication $\sigma_{\text{high}}$ & 0.25 \\
Code length limit & 2000 chars \\
Length of args\_list $N$ & 4 \\
Number of in-context examples $N_{\text{e}}$ & 2 \\
Priority queue size $M$ & 50 \\
\bottomrule
\end{tabular}
\end{table}

\subsection{Training Infrastructure}

Our implementation is based on the VERL~\cite{sheng2024hybridflow} reinforcement learning framework with vLLM~\cite{kwon2023efficient} for efficient inference. The training pipeline consists of:

\begin{itemize}
\item \textbf{Challenger training}: Uses 4 A100 GPUs with FSDP (Fully Sharded Data Parallel) strategy. The Solver model runs as a vLLM service on 4 separate GPUs to evaluate question difficulty in real-time during rollout.

\item \textbf{Solver training}: Uses 8 A100 GPUs with FSDP strategy.
\end{itemize}

\subsection{Seed Code Examples}

The code example queue $Q$ is initialized with four seed examples, as shown in Figures~\ref{fig:seed_jigsaw}-\ref{fig:seed_bbox}:
(1)~\textbf{Jigsaw Puzzle} (Figure~\ref{fig:seed_jigsaw}): divides the image into a $2 \times 2$ grid and permutes blocks according to a specified order;
(2)~\textbf{Rotation} (Figure~\ref{fig:seed_rotation}): rotates the image by a specified angle with bicubic resampling;
(3)~\textbf{Cropping} (Figure~\ref{fig:seed_cropping}): extracts different regions defined by normalized bounding box coordinates;
(4)~\textbf{Bounding Box Drawing} (Figure~\ref{fig:seed_bbox}): draws red rectangles on the image at specified coordinates.

\begin{figure*}
\begin{promptbox}[breakable=false]{Jigsaw Puzzle}
from PIL import Image

def edit_image(img: Image.Image, N: int, order: list) -> Image.Image:
    width, height = img.size
    block_w = width // N
    block_h = height // N
    adjusted_size = (N * block_w, N * block_h)
    img = img.resize(adjusted_size)
    out_img = Image.new("RGB", adjusted_size)
    for new_idx in range(N * N):
        new_row = new_idx // N
        new_col = new_idx % N
        orig_idx = order[new_idx]
        orig_row = orig_idx // N
        orig_col = orig_idx % N
        left = orig_col * block_w
        upper = orig_row * block_h
        right = left + block_w
        lower = upper + block_h
        block = img.crop((left, upper, right, lower))
        out_img.paste(block, (new_col * block_w, new_row * block_h))
    return out_img

args_list = [
    {'N': 2, 'order': [3, 0, 1, 2]},
    {'N': 2, 'order': [1, 0, 3, 2]},
    {'N': 2, 'order': [0, 3, 2, 1]},
    {'N': 2, 'order': [2, 1, 0, 3]},
]
\end{promptbox}
\vspace{-2ex}
\caption{Seed code example: Jigsaw Puzzle.}
\label{fig:seed_jigsaw}
\end{figure*}

\begin{figure*}
\begin{promptbox}[breakable=false]{Rotation}
from PIL import Image

def edit_image(img: Image.Image, angle: float) -> Image.Image:
    return img.rotate(angle, expand=True, resample=Image.Resampling.BICUBIC)

args_list = [
    {'angle': 15},
    {'angle': 45},
    {'angle': 90},
    {'angle': 180},
]
\end{promptbox}
\vspace{-2ex}
\caption{Seed code example: Rotation.}
\label{fig:seed_rotation}
\end{figure*}

\begin{figure*}
\begin{promptbox}[breakable=false]{Cropping}
from PIL import Image

def edit_image(img: Image.Image, bbox_2d: list) -> Image.Image:
    real_width, real_height = img.size
    bbox_2d = [int(b * real_width / 1000) if i % 2 == 0 else int(b * real_height / 1000) for i, b in enumerate(bbox_2d)]
    return img.crop(bbox_2d)

args_list = [
    {"bbox_2d": [205, 220, 335, 422]},
    {"bbox_2d": [103, 94, 378, 210]},
    {"bbox_2d": [452, 603, 565, 750]},
    {"bbox_2d": [154, 752, 357, 958]},
]
\end{promptbox}
\vspace{-2ex}
\caption{Seed code example: Cropping.}
\label{fig:seed_cropping}
\end{figure*}

\begin{figure*}
\begin{promptbox}[breakable=false]{Bounding Box Drawing}
from PIL import Image, ImageDraw

def edit_image(img: Image.Image, bbox_2d: list) -> Image.Image:
    draw = ImageDraw.Draw(img)
    bbox_2d = [int(b * real_width / 1000) if i % 2 == 0 else int(b * real_height / 1000) for i, b in enumerate(bbox_2d)]
    draw.rectangle(bbox_2d, outline="red", width=5)
    return img

args_list = [
    {"bbox_2d": [205, 220, 335, 422]},
    {"bbox_2d": [103, 94, 378, 510]},
    {"bbox_2d": [452, 610, 556, 850]},
    {"bbox_2d": [154, 750, 357, 958]},
]
\end{promptbox}
\vspace{-2ex}
\caption{Seed code example: Bounding Box Drawing.}
\label{fig:seed_bbox}
\end{figure*}

\subsection{Templates}

We present the prompt templates used in our framework. The Challenger prompt (Figure~\ref{fig:tpl_challenger}) instructs the model to generate an executable \texttt{edit\_image} function along with four parameter sets, using two in-context examples sampled from the priority queue $Q$. The Solver prompt (Figure~\ref{fig:tpl_solver}) is intentionally minimal, presenting the VQA question and requesting the answer in a boxed format for easy extraction. The two VQA synthesis templates (Figures~\ref{fig:tpl_p2i} and~\ref{fig:tpl_i2p}) construct the two types of questions from the Challenger's code execution results: Parameter-to-Image asks the Solver to identify which output image corresponds to a given parameter set, while Image-to-Parameter requires the Solver to determine which parameters produced a specific output image.

\begin{figure*}[!ht]
\begin{promptbox}[breakable=false]{Prompt Template for Challenger}
# Task
Write a simple Python function named `edit_image` that edit the user's image, and then design 4 different parameter sets.

# Requirements
1. The `edit_image` function must accept a PIL Image object and specific parameters, returning a modified PIL Image object.
2. The Python code must include necessary imports, the `edit_image` function, and a list of dictionaries named `args_list`.
3. Ensure that the 4 sets of parameters in `args_list` produce 4 visually distinct editing results.
4. No comments must be added to the `edit_image` function.
5. The examples below are for format reference only. The parameters must be designed according to the specific content of the user's image; do not copy them directly.

# Examples

## Example 1
```python
{code_str1}
```

## Example 2
```python
{code_str2}
```

Observe the given image, design the code with reference to the image content. Output your reasoning process inside <thinking>...</thinking> tags, followed by the final Python code in
```python
...
```
\end{promptbox}
\vspace{-2ex}
\caption{Prompt template for the Challenger.}
\label{fig:tpl_challenger}
\end{figure*}

\begin{figure*}[!ht]
\begin{promptbox}[breakable=false]{Prompt Template for Solver}
{question}
Please put your final answer within \\boxed{}.
\end{promptbox}
\vspace{-2ex}
\caption{Prompt template for the Solver.}
\label{fig:tpl_solver}
\end{figure*}

\begin{figure*}[!ht]
\begin{promptbox}[breakable=false]{Template for VQA Synthesis (Parameter-to-Image)}
The given images are image_0, image_1, image_2, image_3, and image_4, respectively. Images image_1 through image_4 are the results of applying the `edit_image` function to image_0 with different arguments.
```python
{code_str}
```
Question: After applying the `edit_image` function to image_0 with `{arg_chosen}`, which candidate image will be produced?
Options:
A. image_1
B. image_2
C. image_3
D. image_4

Guidelines: First, understand the the code, then carefully observe the images, and solve the problem based on the key elements or details in the images.
\end{promptbox}
\vspace{-2ex}
\caption{VQA synthesis template for Parameter-to-Image questions.}
\label{fig:tpl_p2i}
\end{figure*}

\begin{figure*}[!ht]
\begin{promptbox}[breakable=false]{Template for VQA Synthesis (Image-to-Parameter)}
The given images are image_0, image_1, image_2, image_3, and image_4, respectively. Images image_1 through image_4 are the results of applying the `edit_image` function to image_0 with different arguments.
```python
{code_str}
```
Question: Which set of arguments, when passed to `edit_image` for image_0, produces image_1?
Options:
A. {arg_1}
B. {arg_2}
C. {arg_3}
D. {arg_4}

Guidelines: First, understand the the code, then carefully observe the images, and solve the problem based on the key elements or details in the images.
\end{promptbox}
\vspace{-2ex}
\caption{VQA synthesis template for Image-to-Parameter questions.}
\label{fig:tpl_i2p}
\end{figure*}

\section{Additional Experimental Results}
\label{sec:appendix_results}

\subsection{Qualitative Examples}

\begin{figure*}
\centering
\includegraphics[width=\linewidth]{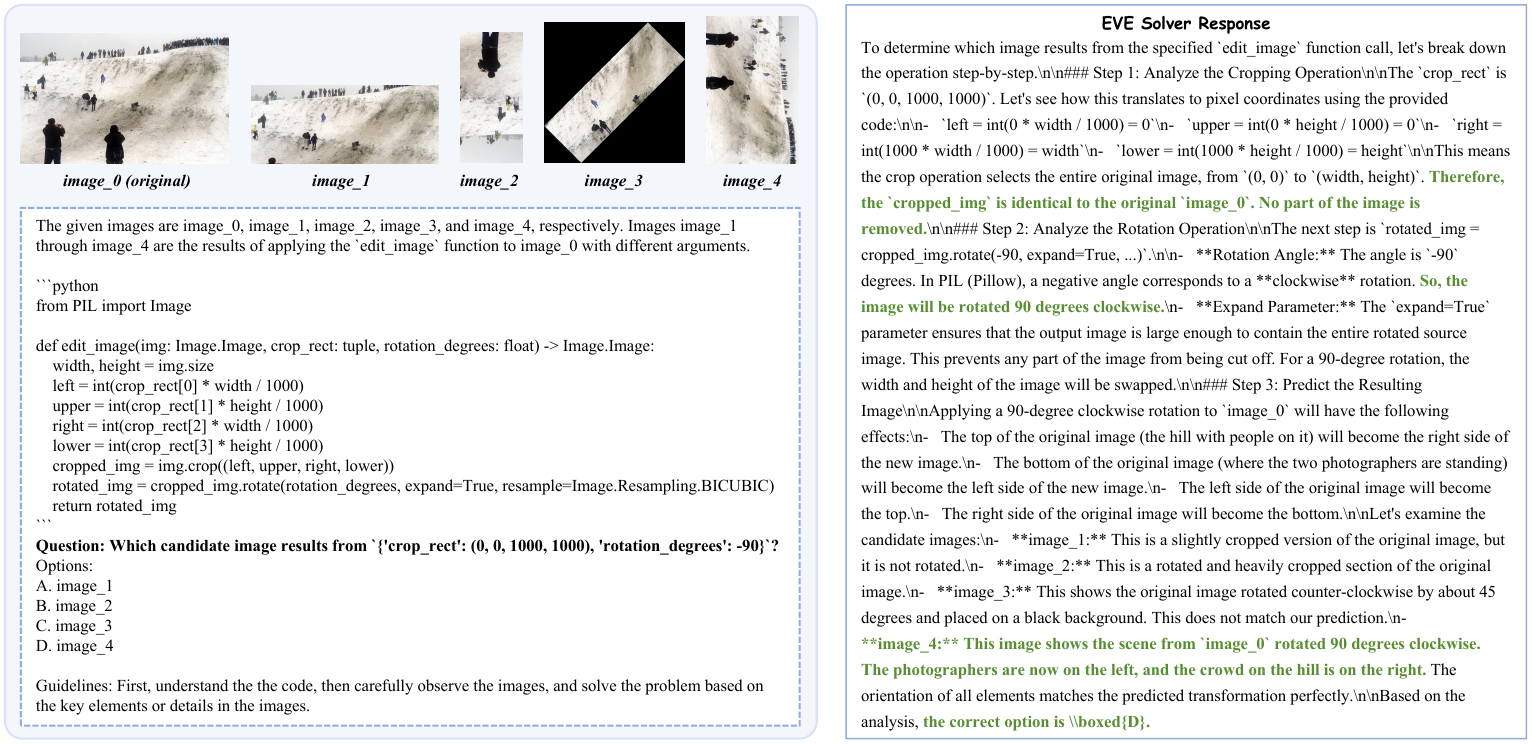}
\vspace{-2ex}
\caption{A Parameter-to-Image example. Left: the synthesized question with code, parameters, and candidate images. Right: the EVE Solver's step-by-step reasoning process leading to the correct answer.}
\label{fig:case_p2i}
\end{figure*}

\begin{figure*}
\centering
\includegraphics[width=\linewidth]{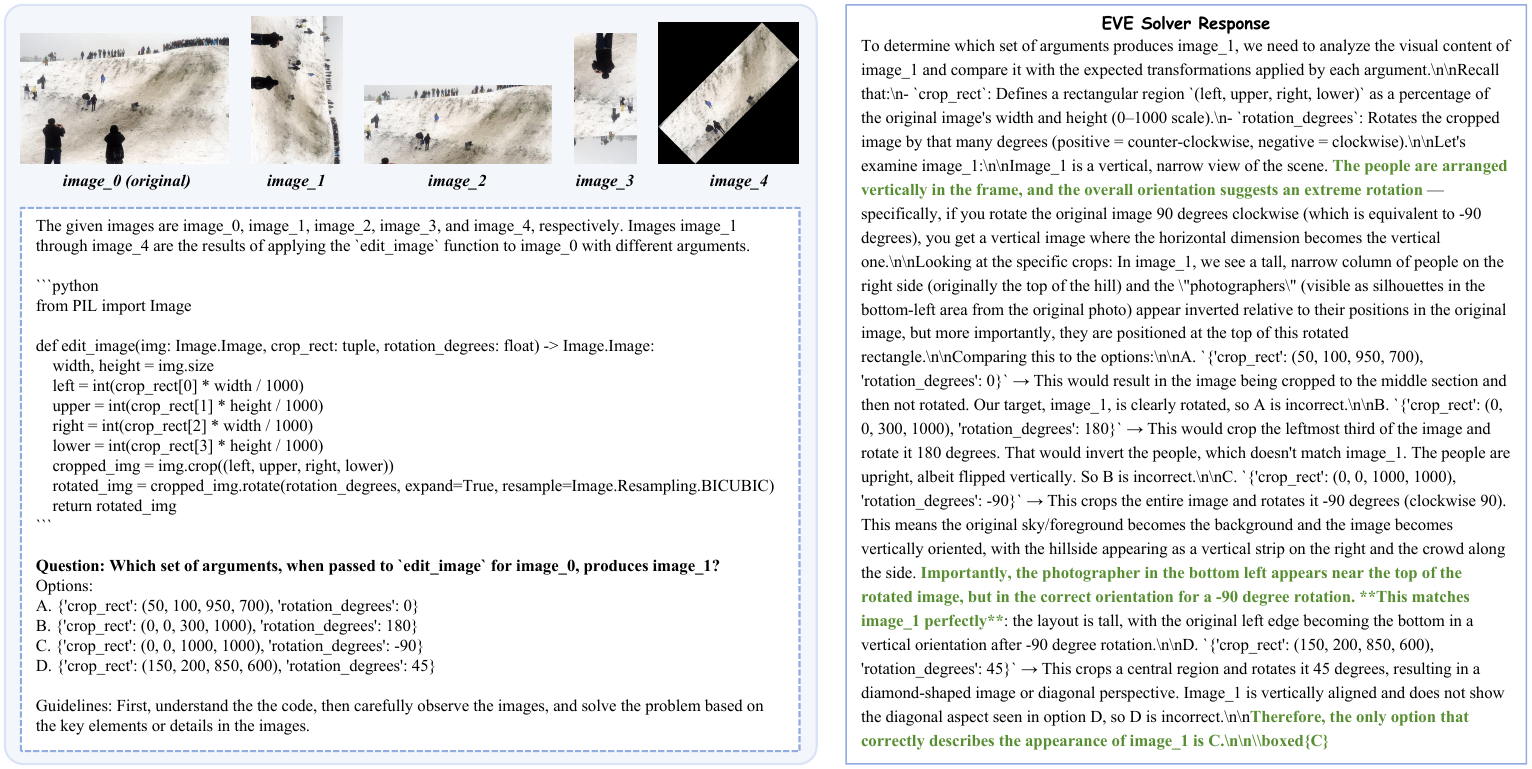}
\vspace{-2ex}
\caption{An Image-to-Parameter example. Left: the synthesized question with code, candidate parameters, and the target output image. Right: the EVE Solver's reasoning process that correctly identifies the parameters.}
\label{fig:case_i2p}
\end{figure*}

\begin{figure*}
\centering
\includegraphics[width=\linewidth]{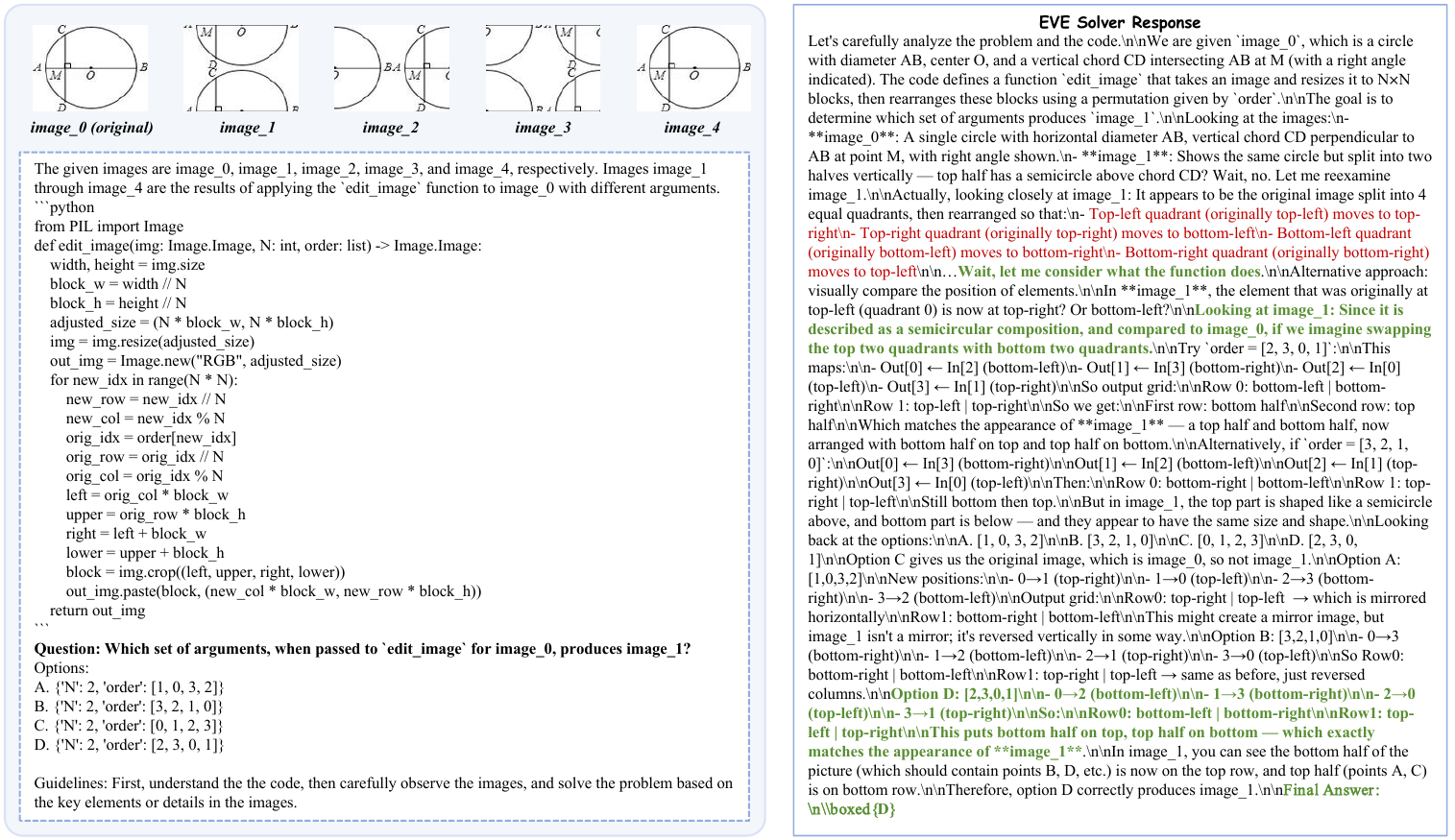}
\vspace{-2ex}
\caption{A Jigsaw Puzzle example (Image-to-Parameter) demonstrating multi-step reasoning and self-correction. \textcolor{red}{Red} highlights indicate reasoning errors that the model subsequently identifies and corrects (\textcolor{green!60!black}{green} highlights), ultimately reaching the correct answer through reflection.}
\label{fig:case_reflect}
\end{figure*}

We present qualitative examples in Figures~\ref{fig:case_p2i}-\ref{fig:case_reflect} to illustrate how the EVE Solver reasons through the synthesized VQA tasks. Each example shows the complete question on the left and the Solver's full reasoning process on the right.

\section{Limitations and Future Directions.}
While EVE transfers to higher-level reasoning tasks, 2D pixel operations have inherent limitations in encoding physical commonsense or complex compositional relationships. Future work can integrate richer executable environments—3D rendering engines for physical reasoning, or UI rendering code for hierarchical layout understanding. Performance plateaus after 4-5 iterations suggest that continued improvement may require more sophisticated curriculum strategies or larger base models. Extending to other modalities (video, 3D, audio) also remains promising.

\end{document}